\documentclass[journal]{IEEEtran}
\usepackage{amsmath,amsfonts}
\usepackage{algorithmic}
\usepackage{algorithm}
\usepackage{array}
\usepackage[caption=false,font=normalsize,labelfont=sf,textfont=sf]{subfig}
\usepackage{textcomp}
\usepackage{stfloats}
\usepackage{url}
\usepackage{verbatim}
\usepackage{graphicx}
\usepackage{multirow}
\usepackage{booktabs}
\usepackage{xcolor}
\usepackage{colortbl}
\usepackage{cite}
\usepackage{tablefootnote}
\usepackage{color}
\usepackage[colorlinks, 
citecolor=green,
linkcolor=red,
urlcolor=blue]{hyperref}

\begin{document}

\renewcommand{\arraystretch}{1.2}

\title{Sat-DN: Implicit Surface Reconstruction from Multi-View Satellite Images with Depth and Normal Supervision}

\author{Tianle Liu, Shuangming Zhao\textsuperscript{$\dagger$}, Wanshou Jiang\textsuperscript{$\dagger$}, Bingxuan Guo
\thanks{Tianle Liu and Shuangming Zhao are with the School of Remote Sensing Information Engineering, Wuhan University, Wuhan, China (email: tianleliu@whu.edu.cn; smzhao@whu.edu.cn).}
\thanks{Wanshou Jiang and Bingxuan Guo are with the State Key Laboratory of Information Engineering in Surveying, Mapping and Remote Sensing, Wuhan University, Wuhan, China (email: jws@whu.edu.cn; b.guo@whu.edu.cn).}
\thanks{$\dagger$ indicates the corresponding author.}
}

\markboth{Journal of \LaTeX\ Class Files,~Vol.~14, No.~8, August~2021}%
{Shell \MakeLowercase{\textit{et al.}}: A Sample Article Using IEEEtran.cls for IEEE Journals}


\maketitle

\begin{abstract}
With the advancement of satellite imaging technology, acquiring high-resolution multi-view satellite imagery has become increasingly accessible, enabling efficient and location-independent ground model reconstruction. However, traditional stereo matching methods often miss fine details, while neural radiance fields (NeRFs) provide high-quality reconstructions at the cost of prohibitively long training times. Moreover, challenges such as low visibility of building facades, illumination inconsistencies, and weakly textured regions further hinder accurate geometry recovery. To tackle these issues, we propose \textit{Sat-DN}, a novel depth- and normal-guided surface reconstruction framework for satellite imagery. Given multi-view satellite images and predicted depth maps, Sat-DN employs a progressively trained multi-resolution hash grid architecture to efficiently represent scene geometry. Our pipeline integrates explicit depth supervision and surface normal consistency constraints to preserve structural boundaries and planar regularities. The progressive training strategy first fits low-frequency geometric structures and then refines high-frequency details under the guidance of depth and normals. Extensive experiments on the DFC2019 dataset demonstrate that Sat-DN significantly outperforms traditional MVS and NeRF-based baselines in both accuracy and efficiency. The code is publicly available at \url{https://github.com/costune/SatDN}.
\end{abstract}

\begin{IEEEkeywords}
 Neural surface reconstruction, Neural radiance field, Satellite image, Volume rendering.
\end{IEEEkeywords}

\section{Introduction}
\IEEEPARstart{R}{econstructing} three-dimensional surfaces from satellite images present significant challenges in the field of remote sensing and computer vision\cite{zhao2023review}. Using the wide coverage and high resolution of satellite images, a rapid large-scale reconstruction of regions can be achieved, which plays a crucial role in urban planning and digital twin applications\cite{facciolo2017automatic,Leotta_2019_CVPR_Workshops,zhang_leveraging_2019,mari2022sat}.

In recent years, approaches leveraging neural radiance fields (NeRFs)\cite{mildenhall2021nerf,barron2021mipnerf} have demonstrated promising results in synthesizing photorealistic views of complex scenes. Neural radiance fields encode a scene as an implicit neural field and utilize differentiable volume rendering\cite{mildenhall2021nerf} techniques to render the field into images. Compared with multi-view ground truth images, it can optimize the density and radiation distribution of the scene through gradient propagation. However, directly applying NeRFs to satellite images is not a trivial thing. High-resolution satellite images in high dynamic range (HDR) are described by the Rational Polynomial Coefficients (RPC), which differ from the ordinary intrinsic and extrinsic coefficients determined images. In addition, limited perspectives, huge illumination variances, and extensive weakly textured regions hinder the achievement of better reconstruction results.

\begin{figure}[t]
    \centering
    \includegraphics[width=1\linewidth]{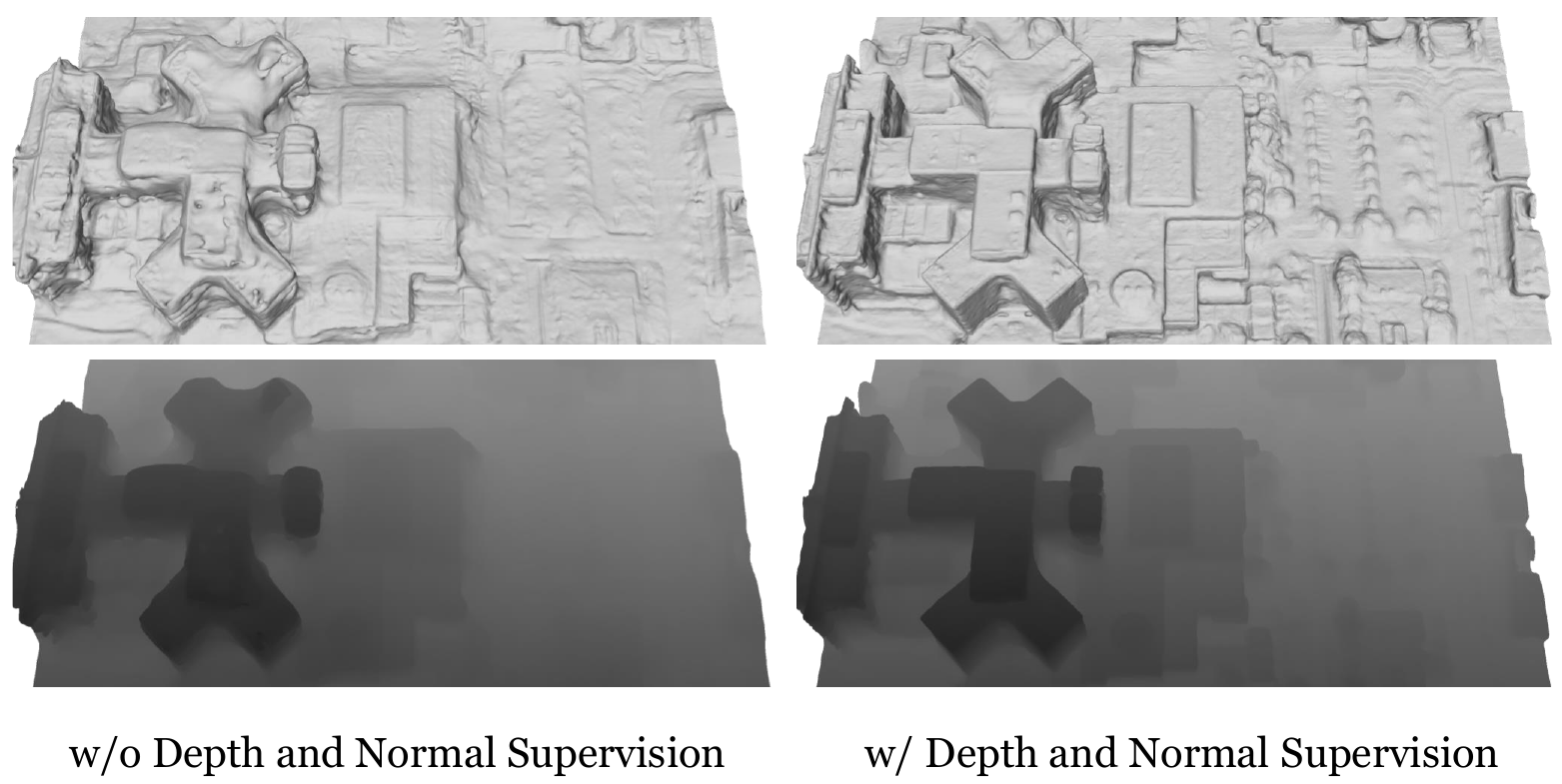}
    \caption{The effects of our proposed depth and normal supervision. We utilize multi-view satellite imagery combined with explicit depth and surface normal consistency supervision to reconstruct detailed surface models of the terrain.}
    \label{fig:first}
\end{figure}

To address these issues, some previous works have already achieved remarkable results\cite{zhang_leveraging_2019,derksen2021shadow,mari2022sat,Mari_2023_eo,zhang_fvmd-isre_2024}. The common NeRFs pipeline is tailored for images defined by intrinsic and extrinsic parameters. Zhang \textit{et al.}\cite{zhang_leveraging_2019} samples sparse points by the RPC model and optimizes the satellite image camera parameters through bundle adjustment with COLMAP\cite{schoenberger2016sfm}, which bridges the gap between the RPC model and matrix parameters. However, the translation matrix obtained has a relatively large eigenvalue, which can lead to numerical instability in computations, and the camera matrix representation is not suitable for subsequent evaluations. S-NeRF\cite{derksen2021shadow} is the first attempt to construct neural radiance fields using satellite data. It achieves spatial point sampling through ground elevation and introduces an additional multilayer perceptron (MLP) to predict satellite shadows, rather than explicitly decomposing the lighting and geometry. Sat-NeRF\cite{mari2022sat} introduces an uncertainty module to address inconsistencies caused by dynamic vehicles. It also samples Rational Polynomial Coefficients to model the ray projection process, enabling the construction of an RPC-based neural radiance field. However, multiple queries of the MLP structure result in significant computational overhead. FVMD-ISRe\cite{zhang_fvmd-isre_2024} \textit{et al.} integrates the RPC ray projection process into NeuS\cite{wang2021neus}, enabling fine reconstruction with few number of views. However, the method relies on manually selecting reconstruction perspectives and cannot address the issue of missing views around the terrain.

In this paper, we aim to develop a method for end-to-end surface reconstruction, mitigating the impact of satellite image characteristics on reconstruction quality and overcoming the limitations of existing methods. The challenges in surface reconstruction from satellite images include: \textbf{\textit{First}}, the color of pixels on different images corresponding to the same spatial location differ significantly, and the reflection intensity varies dramatically between areas on one image. Due to the limitations of satellite imaging technology, multi-view images of a single geographic location require multiple satellite passes to capture, resulting in multi-temporal characteristics among images from one set. Because of the sunlight angle, the building rooftop areas experience strong reflective glare, causing overexposure in the color transitions of these regions. The inconsistency in lighting and shading causes unstable optimization, which leads to geometric inaccuracy\cite{martin2021nerfw,rematas2022urban,Sabour_2023_robustnerf}. \textbf{\textit{Second}}, satellite images contain extensive repeated textures and weakly textured regions, which result in the reconstructed surface being insufficiently smooth, and the normals lacking global consistency. This leads to irregular fluctuations of the plane\cite{wang2021neus,Fu2022GeoNeus,Wang2022NeuS2}, resulting in a reduction of reconstruction results. \textbf{\textit{Third}}, Due to the limited number of satellite image viewpoints and constrained tilt angles, occlusion often occurs in the images of the reconstruction areas, and visibility of information such as building facades is reduced. This can lead to errors in reconstructing complex building geometries and cause the model to produce holes\cite{kangle2021dsnerf,guo2023streetsurf}.

We address these problems one by one. \textbf{\textit{First}}, we propose to leverage monocular depth cues from a pre-trained depth estimation model\cite{depth_anything_v2} as an optimization constraint. We adjust the scale and offset of monocular relative depth using a point cloud refined by triangulation and bundle adjustment. Due to the smoothness of the predicted depth values, the disturbance of the illumination variation in the images on the depth values is minimal. We can use adjusted depth information to guide geometry, avoiding ambiguities in photometric optimization. \textbf{\textit{Second}}, we fuse depth values to obtain monocular normal information and derive angular similarity from the normals. Non-edge pixels are treated as planar surfaces, and we apply a surface normal consistency constraint to guide the smooth distribution of normals, which helps achieve better results in large planar areas such as rooftops and roads. \textbf{\textit{Third}}, considering the smoothness inherent attribute of interpolation, we introduce a multi-res hash feature grid\cite{mueller2022instant} to store and optimize scene features, effectively reducing inference time. Additionally, we employ a progressive training strategy to prevent holes and indentations when reconstructing building facades.

In summary, we present a new surface reconstruction pipeline from satellite images, called \textbf{Sat-DN}, coupled with the multi-res hash feature\cite{mueller2022instant} grid and neural surface reconstruction\cite{wang2021neus} using explicit monocular cues guidance. We evaluate our pipeline on the widely-used DFC2019\cite{dfc2019data} dataset. The efficiency is shown in Fig. \ref{fig:first}. Quantitative and qualitative experiments indicate that our method outperforms other surface reconstruction methods in the accuracy of reconstructed mesh and digital surface model (DSM). Our contributions are included as follows:

\begin{enumerate}
    \item We introduce monocular depth supervision and surface normal consistency supervision into satellite image surface reconstruction to mitigate the impact of lighting differences and weak or repeated textures on reconstruction accuracy.
    \item We integrate the multi-res hash feature grid into the pipeline to accelerate the training. A progressive coarse-to-fine training strategy is also introduced to prevent the occurrence of holes and indentations in the reconstructed model.
    \item By conducting experiments on Jacksonville and Omaha from the DFC2019\cite{dfc2019data} dataset, Sat-DN outperforms the SOTA methods, demonstrating the effectiveness of our proposed method.
\end{enumerate}

The subsequent part of the paper is structured as follows. Section \ref{related_works} reviews the 3D reconstruction of satellite images and neural implicit representations. Section \ref{preliminary} details the preliminaries utilized in the proposed method. Section \ref{method} provides a comprehensive explanation of the structure and methodology adopted in Sat-DN. Section \ref{experiments} presents the results of Sat-DN on the dataset and comparisons with other state-of-the-art (SOTA) methods. In section \ref{conclusion}, we conclude the proposed method.

\section{Related Works}\label{related_works}
In this section, we review traditional 3D reconstruction methods from satellite images relevant to our research and emerging methods that leverage volume rendering for reconstruction.

\subsection{3D Reconstruction Using Satellite Images}
Reconstructing 3D surface models from satellite images has been a longstanding challenge\cite{duan2016towards,facciolo2017automatic,zhao2021double,stucker2022resdepth,gao2023general}. Traditional reconstruction pipelines, mainly dominated by stereo matching, focus on pair-view reconstruction, consisting of four steps: stereo rectification, stereo matching, triangulation, and multi-view DSM fusion\cite{zhao2023review}. Stereo matching or disparity estimation is the process of finding the pixels in the different views that correspond to the same 3D point in the scene\cite{OROZCO2016121}. As a representative of global stereo matching (GSM), Kolmogorov and Zabih\cite{kolmogorov2001computing} introduced an energy minimization formulation of the correspondence problem with occlusions and a fast graph-cut-based approximation algorithm. However, GSM is not suitable for processing large-scale satellite images uniformly, which brings considerable computation burden. Semi-global stereo matching (SGM)\cite{hirschmuller2005accurate,hirschmuller2007stereo} uses an efficient one-dimensional path aggregation method to compute the minimum of the energy function, replacing the two-dimensional minimization algorithm used in GSM. Furthermore, the variations of SGM further improved accuracy and efficiency, such as MGM\cite{Facciolo2015MGMAS}, SGBM\cite{lastilla2019}. Ghuffar\cite{ghuffar2016satellite} applied the SGM to generate DSM from satellite image space and georeferenced voxel space avoiding stereo rectification and triangulation. Dumas \textit{et al.}\cite{dumas2022improving} combined semantic segmentation with SGM, using semantic guidance for pixels to stop matching paths at semantic boundaries, effectively optimizing the disparity transition between buildings and other areas. Yang \textit{et al.}\cite{yang2020novel} proposed semi-global and block matching, using adaptive block matching instead of the dense matching strategy. Barnes \textit{et al.}\cite{barnes2009patchmatch} proposed the PatchMatch method, which created a support window for every pixel in the reference image and used the slide window method to calculate the matching cost. He \textit{et al.}\cite{he2019learing} introduced a multi-support-patches extraction block to extract multispectral central-surround information. Bleyer \textit{et al.}\cite{bleyer2011patchmatch} improved the fronto-parallel window to slanted support window, which extended the matching accuracy to the sub-pixel level. Notably, S2P\cite{franchis2014pipeline,franchis2014refine,franchis2014rect} plays an important role in satellite image matching as a classical method. It approximated satellite images using a pinhole camera model, allowing the application of traditional reconstruction pipelines. Modern deep learning methods have brought stereo matching into a new stage. PSMNet\cite{chang2018pyramid} used spatial pyramid pooling and 3D CNN to aggregate context and regularize cost volume. Yang \textit{et al.}\cite{yang2019hsm} proposed a hierarchical stereo matching architecture to achieve high-res matching through a coarse-to-fine strategy. HMSM-Net\cite{he2022hmsmnet} utilized an end-to-end disparity learning model using low-level spatial features to guide the cost volume fusion of high-level features, and achieved good results in occluded and repeated areas. Sat-MVSF\cite{gao2023general} proposed a uniform framework for multi-view stereo (MVS) including pre-processing, an MVS network, and post-processing, which can generalize in different images.

Although traditional stereo matching reconstruction methods are well-established, they still involve complex workflows and require extensive manual intervention. Additionally, factors like lighting, style, and other image parameters continue to hinder the reconstruction quality of traditional approaches. These methods often rely on stereo image pairs to recover DSM and suffer from insufficient geometric accuracy. Multi-view information should be considered to achieve finer model representations.

\subsection{Neural Implicit Representations}
Neural implicit representations have garnered significant attention in recent years for 3D space representation and reconstruction\cite{mildenhall2021nerf}. The core idea of this approach is to use neural networks as continuous functions, embedding geometric and appearance information implicitly into a high-dimensional representation space. Compared to traditional discrete grid-based 3D representations, implicit representations can continuously describe scenes, offering fine details in higher resolution.

In novel view synthesis, NeRF\cite{mildenhall2021nerf} has demonstrated impressive results by introducing positional encoding, which maps three-dimensional space into the frequency domain, enhancing spatial representation capabilities. Mip-NeRF\cite{barron2021mipnerf} further employed integrated positional encoding and modelling the frequency domain projection of a spatial Gaussian distribution at the sampling positions through the view frustum that intersects the pixels. This enhancement increases the sensitivity to depth and minimises distortion. NeRF++\cite{kaizhang2020} models the scene as a unit sphere and applies inverted sphere parameterization to coordinates outside the unit sphere. NeRF-W\cite{martin2021nerfw} introduced feature vectors into the MLP to learn the colour styles of training images and incorporates weighting information, with a focus on learning static objects. URF\cite{rematas2022urban} introduced learnable scene exposure vectors and computes the final synthesized color based on affine transformations. To accelerate training, \cite{yu2022plenoxel,mueller2022instant,sun2022dvgo} introduces a voxel-grid hybrid structure to explicitly represent 3D information, and the features of sampled points are obtained through interpolation. However, the volumetric density in the radiance field cannot be perfectly aligned with the surface, making it challenging to accurately represent geometry.

In 3D reconstruction, surface rendering methods such as IDR\cite{yariv2020multiview} and DVR\cite{Niemeyer2019DifferentiableVR} define radiance directly on the surface and use implicit gradients for rendering. However, these approaches require input pixel-aligned masks. Volume rendering methods like \cite{Oechsle2021unisurf,yariv2021volsdf,wang2021neus}, directly using the photometric as supervision, can reconstruct accurate geometry. UNISURF\cite{Oechsle2021unisurf} derived surface positions of objects based on the zero-level set of SDF, generating realistic views by rendering colors and normals of points intersecting rays with the surface. VolSDF\cite{yariv2021volsdf} assumed maximum volumetric density at the surface, linking volumetric density and SDF values and using an approximation method for sampling. NeuS\cite{wang2021neus} addressed the offset issue in the conversion between volumetric density and SDF, proposing an unbiased transformation function that aligns volumetric density at its maximum with the surface. Sun \textit{et al.}\cite{Sun2022Neural3R} proposed a hybrid voxel representation and surface-guided sampling strategy, accurately reconstructing geometry by constraining the sampling range. NeuS2\cite{Wang2022NeuS2} combines NGP\cite{mueller2022instant} with NeuS to achieve geometric reconstruction of dynamic human bodies. Neuralangelo\cite{li2023neuralangelo} improves the issue of local interpolation grids lacking global smoothness by using numerical gradients to compute higher-order interpolation derivatives. Moreover, Voxurf\cite{wu2022voxurf} proposed a voxel-based 3D surface reconstruction method, acquiring a faster convergence speed than the pure MLP and hybrid representation methods.

There have already been some attempts to use satellite imagery as training input. S-NeRF\cite{kangle2021dsnerf} incorporated the direction of sunlight into the MLP, aiming to decompose surface albedo and ambient light to reconstruct the true color. Sat-NeRF\cite{mari2022sat} focused on the uncertainty of dynamic objects in the scene, reducing the view synthesis artifacts caused by these dynamic objects. EO-NeRF\cite{Mari_2023_eo} leveraged shadow information for geometric inference, ensuring that the shadows cast by buildings align with the scene's geometry. SUNDIAL\cite{behari2024sundial} precisely estimated lighting through secondary shadow ray transmission and introduced geometric regularization to optimize scene geometry jointly. During training, it simultaneously estimates the sun's direction to enhance the accuracy and robustness of decomposing the scene's physical properties. Sat-Mesh\cite{qu_sat-mesh_2023} leveraged photo consistency constraints to ensure that the textures in synthesized views are consistent with those at the same locations in the training images. This aids the model in learning accurate and plausible geometry. Sat-NGP\cite{billouard2024satngp} was the first to apply hash grids to satellite image reconstruction, but it was limited to novel view synthesis and DSM generation, unable to reconstruct complete surface models. Although the methods mentioned above consider lighting, the limited number and constrained angles of view make reconstructing complex scenes a significant challenge. By directly constraining the geometry, we can achieve better reconstruction results. Thus, we introduce new regularization terms to enhance the reconstruction result.

\section{Preliminary}\label{preliminary}
\subsection{NeRF and NeuS}
Neural radiance fields\cite{mildenhall2021nerf} map each 3D position \(\mathbf{x}=(x,y,z)\in\mathbb{R}^3\) and viewing direction \(\mathbf{d}=(\theta,\phi)\in\mathbb{R}^2\) to the volumetric density \(\sigma\in\mathbb{R}_+\) at that location and the RGB color value \(\mathbf{c}\in\mathbb{R}^3\) emitted from it along the viewing direction to the ray's origin. The color of a ray \( \mathbf{r} \) emitted from the origin \( \mathbf{o} \) in the viewing direction \( \mathbf{d} \) is synthesized through volume rendering by compositing the volumetric density \( \sigma_i \) and color \( \mathbf{c}_i \) of sampled points \(\mathbf{x}_i=\mathbf{o}+t_i\mathbf{d}\) along the ray according to quadrature rule\cite{max1995optical}:
\begin{equation}
\label{equ_1}
    C(\mathbf{r})=\sum_{i=1}^N T_i \alpha_i \mathbf{c}_i,\ \ \ T_i=\prod_{j=1}^{i-1}\left(1-\alpha_i\right),
\end{equation}
where \(T_i\) and \(\alpha_i\) denote accumulated transmittance and alpha value of position \(i\). The alpha value \(\alpha_i=1-e^{-\sigma_i \delta_i}\) and \(\delta_i=t_{i+1}-t_i\) is the distance between adjacent positions. NeRFs\cite{mildenhall2021nerf} overfit 3D scene information from multi-view images and store the information \((\mathbf{c},\sigma)=f_\theta(\mathbf{x},\mathbf{d})\) in an MLP. The neural network is optimized by comparing the differences between rendered and ground truth images.

NeuS\cite{wang2021neus} connects the signed distance function (SDF) with the neural radiance field through an SDF \(f(\mathbf{x})\) to opaque density \(\rho(\mathbf{x})\) transfer function to align the surface with the location where the volumetric density reaches its maximum along a ray. NeuS\cite{wang2021neus} defined the function as:
\begin{equation}
\label{equ_2}
    \alpha_i = \max\left(\frac{\Phi_s(f(\mathbf{x}_i))-\Phi_s(f(\mathbf{x}_{i+1}))}{\Phi_s(f(\mathbf{x}_{i}))} ,0\right),
\end{equation}
where \(\Phi_s(x)\) denotes the Sigmoid function \(\Phi_s(x)=(1+\exp(-sx))^{-1}\) with bandwidth \(s\). With the intrinsic characteristic of occlusion awareness in volume rendering, the photometric constraint adjusts the zero-level set of SDF to coincide with the scene surface accurately. The total loss is:
\begin{equation}
\label{equ_3}
\begin{split}
    \mathcal{L} &= \mathcal{L}_{color} + \lambda\mathcal{L}_{eikonal} \\
                &= \frac{1}{m}\sum_k|C_k-\hat{C}_k| + \frac{\lambda}{mn}\sum_{k,i}(||\nabla f(\mathbf(\mathbf{x}_{k,i})||_2-1)^2,
\end{split}
\end{equation}
where \(C_k\) and \(\hat{C}_k\) denote the ground-truth color and rendered color, \(m\) and \(n\) represent the number of rays and points in each ray, and the \(\mathcal{L}_{eikonal}\) term is to regularize the SDF distribution.

\subsection{Hash Feature Grid}
Since NeRFs\cite{mildenhall2021nerf} use MLPs to encode scene information, multiple MLP queries result in substantial computational overhead, with scene reconstruction training times reaching up to 8-10 hours. M{\"u}ller \textit{et al.} proposed Instant-NGP\cite{mueller2022instant}, which encodes the scene into a multi-res hash grid. The features of a sampled position are concatenated by interpolations of multi-res grid vertex features where the point resides. Specifically, for every sampled point \(\mathbf{x}\in\mathbb{R}^3\), the feature \(h_i(\mathbf{x})\in\mathbb{R}^d\) at each grid level \(i\) is interpolated by the learnable vertex features that looked up in the hash table. After stacking \(L\) different levels' features, the overall mapped feature \(\mathbf{h}(\mathbf{x})=\{h_1(\mathbf{x}),h_2(\mathbf{x}),...,h_L(\mathbf{x})\}\in\mathcal{R}^{L\times d}\) inputs to a tiny MLP to fit SDF value. Because of the implementation of CUDA, the whole process can be computed swiftly without sacrificing geometry and appearance.

\section{Method}\label{method}
\begin{figure*}[!t]
\centering
\includegraphics[width=1\linewidth]{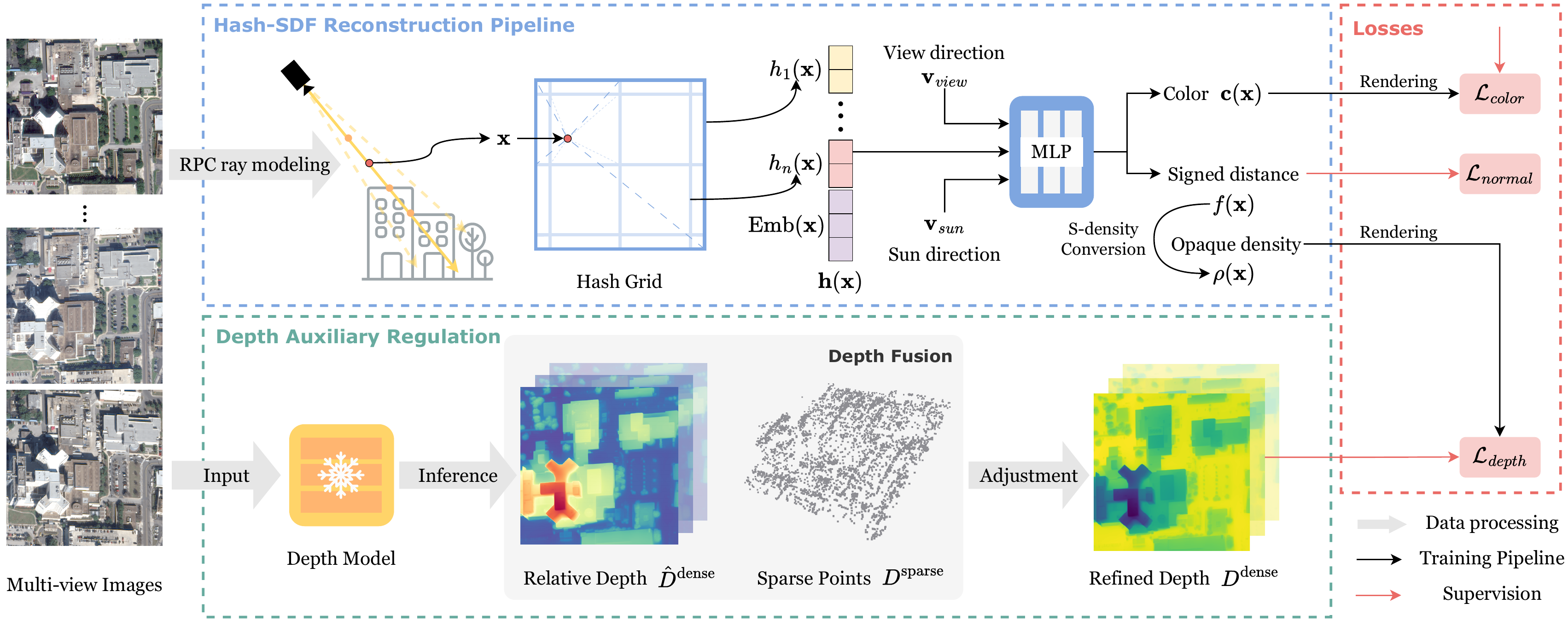}
\caption{Pipeline of our method. Sat-DN takes multi-view satellite imagery as input. The imagery undergoes RPC ray modelling to generate spatial sampling positions, which are then encoded into features using a multi-resolution hash grid. An MLP receives the hash-encoded spatial sampling point positions, frequency-encoded pixel ray directions, and the sun direction as input and outputs the color and SDF value of the spatial sampling position. A pre-trained depth model estimates the relative depth information for the area covered by the satellite images, which is then fused with spatial sparse points obtained through bundle adjustment to generate dense ground-truth depth with absolute scale. By minimizing the differences in pixel color, rendered depth versus dense depth and surface normal angular consistency, the hash features and MLP parameters are optimized. This process enables the end-to-end reconstruction of DSMs and Meshes. The depth and normal loss are further elaborated in Section \ref{depth} and \ref{normal} separately.}
\label{fig:pipeline}
\end{figure*}

Given a set of satellite images \(\{I_i\}_{i=1}^N\), we first model the ray casting process using the RPC model. Next, the sampled points \(\mathbf{x}\) along the rays are input into a multi-res hash grid, where features are interpolated at multiple levels. These features are concatenated with the frequency-embed features of the sampled points \(\text{Emb}(\mathbf{x})\) to form a complete feature vector \(\mathbf{h}(\mathbf{x})\), which is then input into the MLP to obtain both the color \(\mathbf{c}(\mathbf{x})\) and SDF values \(f(\mathbf{x})\) of the sampled points. In addition, relative depth \(\hat{D}_{dense}\) is predicted from the satellite images using a pre-trained depth model. This relative depth is fused with sparse point clouds obtained through triangulation and bundle adjustment of the satellite images, forming a dense depth map \(D_{dense}\) at a real-world scale. The volumetric density distribution is then constrained using depth supervision, optimizing the SDF. Finally, we extract edge information from the depth map, treating non-edge regions as continuous planes and applying smoothness constraints. The following subsections provide a detailed introduction to our proposed method. The pipeline of the method is shown in Fig.\ref{fig:pipeline}.

\subsection{Depth Fusion and Regularization}\label{depth}
Due to the significant variations in lighting conditions in satellite images and the differences in image color styles across multi-temporal data, relying solely on comparing pixel discrepancies between training images and rendered images introduces ambiguities in the optimization process. Therefore, globally consistent depth information is needed to guide the optimization effectively.

Given a satellite image \( I \) and a pre-trained depth model \( F_{\theta} \), we can obtain the relative normalized dense depth of image \( I \) as \(\hat{D}^\text{dense}=F_\theta(I)\). The scale information is calculated from the sparse 3D points \(\mathbf{P} = \{(lon_i, lat_i, alt_i)\in\mathbb{R}^3|i=1,2,...N\}\) obtained through image triangulation and bundle adjustment. Since the 3D points contain only information on longitude, latitude, and altitude, following \cite{mari2022sat}, we obtain sparse depth of a point \(D^\text{sparse}_i\) by calculating the distance between itself and the start point of the ray that goes through the point from the re-parameterized reference plane. Given a 3D sparse point \( P_i\in\mathbf{P} \), whose projection on the 2D image is \( (u_i,v_i)\in\mathbb{R}^2 \), assuming the image coordinate is derived from the corrected satellite RPC model. In addition, we can use RPC to calculate longitude and latitude reversely:
\begin{equation}
\label{equ_4}
    lon_i,lat_i=\text{RPC}(u_i,v_i,alt_i),
\end{equation}
where \(lon_i,lat_i,alt_i\) means longitude, latitude and altitude of point \(P_i\) separately. Since the projection coordinate \((u_i,v_i)\) does not align with the integer pixel, we approximate the projection \((u_i',v_i')=\text{round}(u_i,v_i)\). With the reprojection position \(P_i'=(lon_i',lat_i',alt_i)=(\text{RPC}(u'_i,v'_i,alt_i),alt_i)\) and ray's start point \(P_i^{\text{ref}}=(lon_i^\text{ref},lat_i^\text{ref},alt^\text{ref})=(\text{RPC}(u'_i,v'_i,alt^\text{ref}),alt^\text{ref})\), we can derive the following:
\begin{equation}
\label{equ_5}
        D^\text{sparse}_i=\left\|P_i'-P^\text{ref}_i\right\|_2,
\end{equation}
where \(alt^{\text{ref}}\) means the altitude of the upper reference plane. The entire ray projection process is defined based on the UTM coordinate system, from the reference plane at the highest altitude to the reference plane at the lowest altitude. We treat the sparse points as the true terrain surface positions to correct the dense relative depth. Based on the dense relative depth and sparse absolute depth, we can obtain dense absolute depth as:
\begin{equation}
\label{equ_6}
    D^\text{dense}_i=s\cdot \hat{D}^\text{dense}_i + o,
\end{equation}
where \(\hat{D}^\text{dense}_i\) is relative depth of pixel \((u_i,v_i)\) and \(s\), \(o\) are scale value and offset value, computed through minimizing the differences between sparse absolute depth and fitted depth from the relative one:
\begin{equation}
\label{equ_7}
    s^*,o^*=\arg\min_{s,o}\sum_{i=1}^{HW}|w_i\cdot D^\text{sparse}_i-D^\text{dense}_i(s,o)|^2,
\end{equation}
where \(w_i\in[0,1]\) is a normalized weight representing the reprojection error of the point \(P_i\) onto the image, namely, the fitting confidence for each point. We can use the dense absolute depth \(D^\text{dense}_i=s^*\cdot \hat{D}^\text{dense}_i+o^*\) for pixel depth supervision after calculating \( s^* \) and \( o^* \). However, the depth model trained on a large set of standard photos is not always accurate when applied to satellite images. The depth predicted by the model for water surfaces creates discontinuities with the surrounding land, making depth supervision ineffective. Thus, we apply a semantic mask \(M\) to exclude water regions with erroneous depth in the training stage.

In neural radiance fields, the pixel color is synthesized by the colors of the sampled points along the ray through volume rendering. Similarly, we can obtain the pixel depth using the same method,
\begin{equation}
\label{equ_8}
    \tilde{D}(\mathbf{r})=\sum_{i=1}^NT_i\alpha_id_i,
\end{equation}
where \(d_i\) is the depth of the sampled point \(i\). The \(\alpha_i\) and \(T_i\) of the Eqn. \ref{equ_8} can used the same value from Eqn. \ref{equ_1}, without redundant computation. Depth supervision is applied on a per-pixel basis using the L1 loss:
\begin{equation}
\label{equ_9}
    \mathcal{L}_{depth} = \frac{1}{K} \sum_{i=1}^{K} M_i | \tilde{D}_i - D^\text{gt}_i |,
\end{equation}
where \( \tilde{D}_i \) is the predicted depth and \( D_i^\text{gt} \) is the ground truth dense depth \(D^\text{dense}\) for pixel \( i \). \(M_i\in\{0,1\}\) is the water mask and \(K\) is the batch size. Analogous to an LLFF space, we transform the space between the upper and lower reference planes into a canonical space, which is suitable for optimization.

The depth in our method is reparameterized as the distance between a point on a ray and the ray’s origin on the upper reference plane. We fit the sparse reparameterized depth values from sparse points into a dense relative depth map predicted by the depth model. Due to the large distance between the satellite sensor and the Earth’s surface, and the limited perspective of the satellite, the variation between true depth and reparameterized depth is relatively small. This enables us to obtain an optimal least-squares solution through iterative refinement. To address issues caused by variations in pixel photometry, the depth is used to guide the training rather than to indicate the exact position of the ground-truth surface.

\subsection{Normal Regularization Based on Angular Consistency}\label{normal}
Normals are a critical factor in representing geometry, and a reasonable distribution of normals is essential to ensure geometric accuracy. However, neural radiance fields typically optimize pixel color differences without explicitly constraining geometry in prior methods\cite{kangle2021dsnerf,mari2022sat,qu_sat-mesh_2023,zhang_fvmd-isre_2024,behari2024sundial}. In satellite images, extensively repeated textures or weakly textured areas often exhibit color variations within the same region, meaning that relying solely on color constraints can introduce surface noise, thereby reducing the geometric accuracy of the reconstructed surface model. We introduce a normal constraint based on angular consistency to address this issue.

Due to interference from ground color, directly obtaining normals from satellite images is challenging. We leverage the strong generalization capabilities of depth models, which can provide smooth surface depths and mitigate the impact of color on geometry. Assume the normal for pixel \(i\) obtained from the relative depth map, denoted as the ground truth normal \(\mathbf{n}_i\), is in the pinhole camera coordinate system. We first calculate the angular difference between the normal pixel \(i\) and those of its neighboring pixels:
\begin{equation}
\label{equ_10}
    \delta_i^\text{gt} = \frac{1}{N}\sum_{j\in\text{Adj}(i)}\frac{\mathbf{n}_i\cdot\mathbf{n}_j}{\|\mathbf{n}_i\|\|\mathbf{n}_j\|},
\end{equation}
where \(\text{Adj}(i)\) represents the neighbor {\color{orange}4} pixels around the pixel \(i\), \(N\) is the total number of the neighbors. 

In the SDF network, the normal of the ray \(\mathbf{r}\) is the gradient of the SDF value at sampling point along the ray on the surface \(\mathbf{n}(\mathbf{r})=\nabla f(\mathbf{o}+t\mathbf{d})\), and the resulting normal is in the UTM coordinate system. \(t\) means the distance predicted from the reference plane to the surface. Since the implicit SDF is continuous and gradients approach \(1\), to reduce computational load, we treat the predicted depth of the central pixel as the depth value for nearby pixels and compute the angular difference between ray \(\mathbf{r}\) from pixel \(i\) and neighboring rays from adjacent pixels \(\text{Adj}(i)\) as follows:
\begin{equation}
\label{equ_11}
    \delta_{i}^\text{pred}=\frac{1}{N}\sum_{j\in\text{Adj}(i)}\frac{\nabla f(\mathbf{o}_i+t_i\mathbf{d}_i)\cdot\nabla f(\mathbf{o}_j+t_i\mathbf{d}_j)}{\|\nabla f(\mathbf{o}_i+t_i\mathbf{d}_i)\|\|\nabla f(\mathbf{o}_j+t_i\mathbf{d}_j)\|},
\end{equation}
where \(\mathbf{o}\) and \(\mathbf{d}\) means the origin and direction of the corresponding rays, \(t_i\) is the rendered depth of the ray originated from pixel \(i\). \(N\) represents the number of adjacent pixels as above.

Since the normals calculated from the depth map and the rendered normals are in different coordinate systems, directly constraining the normals themselves is not feasible. Thus, we only constrain the consistency of their distributions. This approach ensures that in planar areas, the angular difference between rendered normals and neighboring normals remains minimal, while in non-planar areas, the difference is greater, guiding the rendered normals closer to the true geometry. Our normal angular consistency loss is defined as follows:
\begin{equation}
    \mathcal{L}_{normal} = \frac{1}{K} \sum_{i=1}^K (\delta_i^\text{pred} - \delta_i^\text{gt})^2,
\end{equation}
where \(i\) means the pixel in the image. By introducing the normal angular consistency loss, we can effectively mitigate geometric ambiguities in color rendering caused by color changes on planar surfaces influenced by the interplay between buildings and sunlight. This reduces the erroneous impact of planar texture colors on geometric optimization.

\subsection{Progressive Training Strategy}
Due to the limited number of training images and constrained viewing angles, the surface model after training often contains holes and discontinuous artifacts beyond the surface contours. We observed that jointly optimizing low-resolution and high-resolution encodings tends to result in surface errors. This is because the model adjusts high-frequency features from the beginning of the train, making it prone to local optimization. Moreover, in satellite imagery, the side facades of individual buildings appear only a limited number of times across all images and often exhibit color inconsistencies, resulting in a lack of consistent color supervision. Therefore, directly optimizing high-frequency details from the beginning is unstable and may lead to the fitting of a multimodal volume density field near the surface. In contrast, optimizing low-frequency features first allows the model to learn a roughly continuous surface distribution during the early stages of training, which helps to prevent significant surface deviations when high-frequency features are subsequently introduced. Therefore, we introduced a progressive training strategy, in which low-resolution grid features are activated first, and higher-resolution features are progressively activated as the training progresses. The feature representation for the sampled point \( \mathbf{x} \) is defined as:
\begin{equation}
    \mathbf{h}(\mathbf{x},\lambda)=\left\{G_1(\lambda)h_1(\mathbf{x}),\dots,G_L(\lambda)h_L(\mathbf{x})\right\},
\end{equation}
where \(G_i(\lambda)\) is a gated function \(\{G_i(\lambda)=1,\lambda\geq i; G_i(\lambda)=0,\lambda<i\}\), \(i\) is the level number and \(h_i(\mathbf{x})\) means the interpolated feature of the input point $\mathbf{x}$ to level \(i\). \( G_i(\lambda) \) can act as a low-pass filter, with low-resolution grids learning the coarse geometry of the scene first, and high-resolution grids learning fine geometry details later. We initialize \( \lambda \) to 4, and it increases \( \lambda \) by 1 every 2.5\% of the total training iterations.

\subsection{Overall Loss Design}
To train Sat-DN, we incorporated a depth loss and a normal angular consistency loss in addition to the original color loss. The objective is to minimize the discrepancies between the rendered and ground-truth images, the rendered and true depths, as well as the angular differences between the pixel normals and their reference values. The overall loss function is defined as:
\begin{equation}
    \mathcal{L} = \mathcal{L}_{color} + \lambda_1\mathcal{L}_{depth} + \lambda_2\mathcal{L}_{normal} + \lambda_3\mathcal{L}_{eikonal},
\end{equation}
where $\lambda_1$, $\lambda_2$ and $\lambda_3$ represent the training weights. Both parameters are defined as $0.1$ in all training scenes.

\section{Experiments}\label{experiments}
In this section, we present the experimental results of Sat-DN. We first explain the dataset used, evaluation metrics, and implementation details of the method. Next, we describe the experimental setup for comparison methods. Finally, we include comparative experiments and ablation studies to demonstrate the effectiveness of our proposed improvements. All experiments were conducted on a system with an Intel i9-14900KF CPU, 64GB of memory, an NVIDIA RTX 4090 GPU running CUDA 11.8, and PyTorch version 2.3.1, using the Ubuntu 22.04 operating system.

\subsection{Datasets and Evaluation Metrics}
\subsubsection{DFC2019}
\begin{figure*}[!ht]
    \centering
    \includegraphics[width=0.9\linewidth]{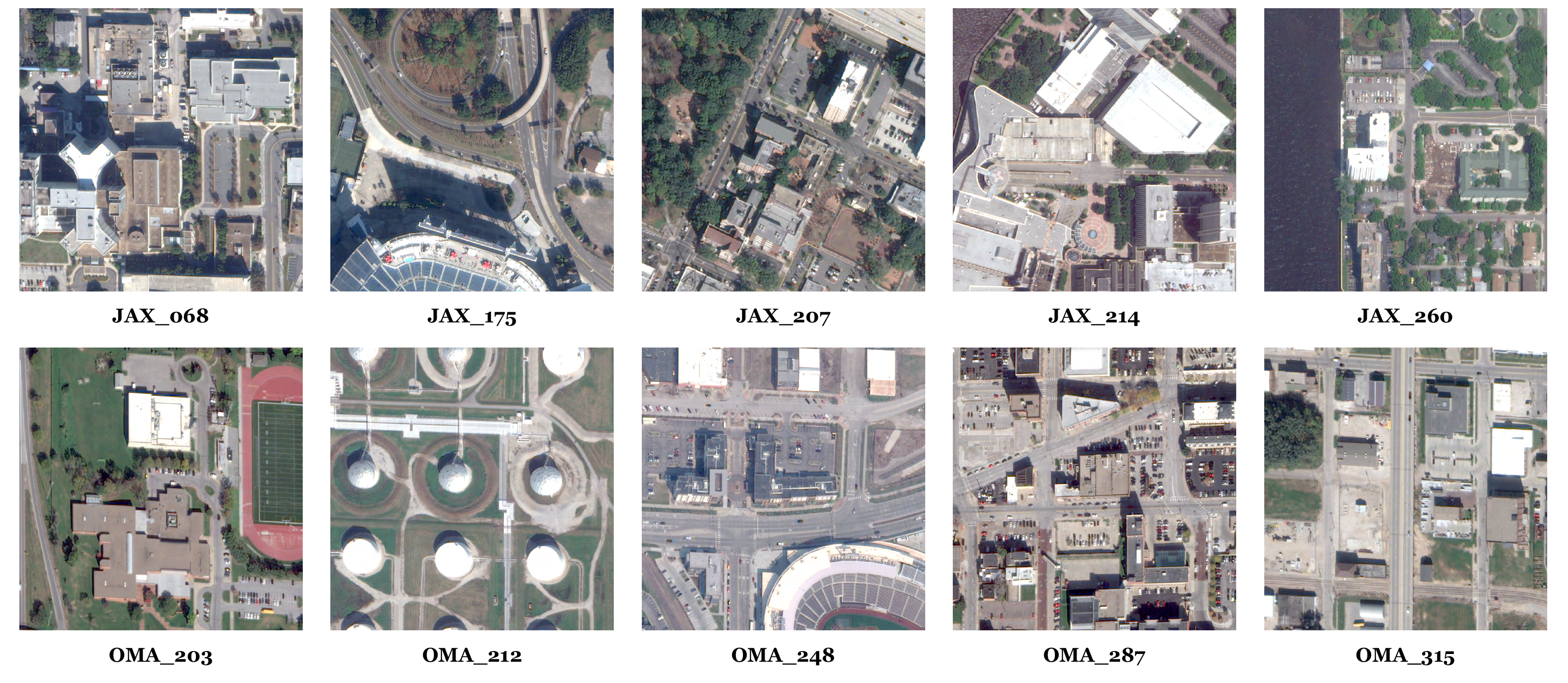}
    \caption{Examples of satellite images from the DFC2019 dataset used in the experiments, covering various geographic locations such as urban areas, forests, and industrial sites.}
    \label{fig:dataset}
\end{figure*}

The DFC2019 (Data Fusion Contest 2019) dataset\cite{dfc2019data}, is a multi-view satellite remote sensing dataset that includes hyperspectral imagery, multispectral imagery, LiDAR point cloud data, and other auxiliary data. The dataset contains satellite images captured over Jacksonville and Omaha, USA, by WorldView-3 over two years. Our task is to reconstruct the 3D surface model of the terrain based on the provided multi-view satellite imagery. The dataset includes DSMs generated from LiDAR point clouds, which serve as ground truth for reconstruction. Each DSM has dimensions of 512$\times$512 pixels with a resolution of 0.5 m, covering an area of 256m $\times$ 256m. To reduce the computational cost, we crop the original satellite images based on the DSM’s RoI and select five scenes from Jacksonville and Omaha for reconstruction. All image RPC parameters were refined through bundle adjustment. Additionally, we use supplementary data\cite{dfc2019supp1,dfc2019supp2} from the DFC2019 dataset to obtain semantic mask information for each image, which is also cropped according to the RoI. To ensure consistency in image style, we refrained from using winter-time satellite images from Omaha, avoiding the potential negative impact of style differences on the reconstruction results. The data parameters we used are shown in Table. \ref{tab_1}, and the scenes with their RoI are illustrated in Fig. \ref{fig:dataset}.

\begin{table}[!ht]
    \centering
    \caption{The information of RoIs used in the experiment.}
    \begin{tabular}{c|cccc}
        \toprule
        \textbf{RoI Name} & \textbf{Images} & \textbf{Anchor (lat, lon)} & \textbf{Zone} & \textbf{RoI Size (m)} \\
        \midrule
        JAX\_068 & 19 & (30.347, -81.665) & 17N & 256$\times$256 \\
        JAX\_175 & 24 & (30.325, -81.638) & 17N & 256$\times$256 \\
        JAX\_207 & 24 & (30.315, -81.681) & 17N & 256$\times$256 \\
        JAX\_214 & 24 & (30.315, -81.665) & 17N & 256$\times$256 \\
        JAX\_260 & 16 & (30.311, -81.665) & 17N & 256$\times$256 \\
        OMA\_203 & 25 & (41.272, -95.952) & 15N & 256$\times$256 \\
        OMA\_212 & 22 & (41.276, -95.921) & 15N & 256$\times$256 \\
        OMA\_248 & 23 & (41.267, -95.934) & 15N & 256$\times$256 \\
        OMA\_287 & 24 & (41.253, -95.941) & 15N & 256$\times$256 \\
        OMA\_315 & 23 & (41.251, -95.855) & 15N & 256$\times$256 \\
        \bottomrule
        \multicolumn{5}{l}{The anchor represents the lat-lon of the RoI's southwestern corner.} \\
        \multicolumn{5}{l}{JAX, OMA represents Jacksonville and Omaha separately.} \\
        \multicolumn{5}{l}{The following number means specified RoI.} \\
    \end{tabular}
    \label{tab_1}
\end{table}

\subsubsection{Evaluation Metrics}
Following the settings of previous works, the evaluation metrics used include MAE (Mean Absolute Error), MED (Median Absolute Error), and CD (Chamfer Distance). MAE calculates the L1 height difference between the reconstructed DSM and the ground truth DSM for each pixel, summing the height differences and then averaging across pixels:
\begin{equation}
    MAE=\frac{1}{N}\sum_{i=1}^N|h^{pred}_i-h^{gt}_i|.
\end{equation}

MED calculates the median of the elevation differences across all pixels:
\begin{equation}
    MED=\text{mediam}\{|h^{pred}_i-h^{gt}_i|,i=1,2,...,N\}.
\end{equation}

Chamfer distance (CD) is a commonly used distance metric to compute the difference between two point sets:
\begin{equation}
    CD\left(S_1, S_2\right)=\frac{1}{S_1} \sum_{x \in S_1} \min _{y \in S_2}\|x-y\|_2^2+\frac{1}{S_2} \sum_{y \in S_2} \min _{x \in S_1}\|y-x\|_2^2.
\end{equation} 

\subsection{Implementation Details}
For the hash grid\cite{mueller2022instant}, we use 24 levels with resolutions ranging from 16 to 2048, with a maximum hash table size of 2\textsuperscript{19} and a feature dimension of 2 per level. The encoding function \(\text{Emd}(.)\) applies a 2\textsuperscript{6} position embedding and concatenates the normalized UTM space coordinates of the sampling points. The SDF MLP consists of 3 layers, taking the hashed features of the sampling point as input and outputting a 1-D SDF value and a 256-D intermediate feature. The color MLP also has 3 layers, with input as the sampling point coordinates, the gradient of the SDF value, the intermediate feature, and the 2\textsuperscript{4} position-embedded ray direction and solar light direction. We follow NeuS\cite{wang2021neus} ray sampling strategy, sampling 128 points per ray. The experiment's batch size is 4096, with 100K iterations, and all of the loss function weights \(\lambda\) are set to 0.1. The normal neighbor number is 4. The output mesh is obtained by marching cubes, and the DSM is generated by the projection of the mesh. Depth-Anything-V2\cite{depth_anything_v2} is used in our method, which is trained on over 62.5M images with high generalization ability.

\subsection{Comparison Methods}
\subsubsection{S2P}
S2P\cite{franchis2014pipeline} uses the MGM method to match ten manually selected stereo image pairs, merging the point clouds generated by each pair to obtain a complete scene point cloud, which is then processed using an medium filter to generate the DSM. The generated DSM matches the parameters of the DSM in the dataset, with other settings following S2P’s default. For missing points in the DSM, interpolation is achieved using inverse distance weighting. Converting the DSM to a mesh involves first transforming the DSM into a point cloud based on elevation and then outputting the mesh from the point cloud using Delaunay triangulation.

\subsubsection{VisSat}
VisSat\cite{VisSat-2019} approximates satellite imagery as a pinhole camera through affine transformations, enabling its use within an MVS pipeline. All experimental settings follow the defaults. After structure-from-motion (SfM), the generated point cloud is projected into ground coordinates to create the DSM, with missing points interpolated using inverse distance weighting. Heights in the DSM are then converted to point clouds, and Delaunay triangulation is applied to obtain the Mesh.

\subsubsection{S-NeRF and Sat-NeRF}
S-NeRF\cite{kangle2021dsnerf} and Sat-NeRF\cite{mari2022sat} employ lighting decoupling to reconstruct ground objects. For experimental consistency and fairness, we use their open-source experimental parameters, adjusting the batch size to 4096 and setting the iteration count to 100K, while keeping other settings unchanged.

\subsubsection{SpS-NeRF}
SpS-NeRF\cite{zhang2023spsnerf} derived a pixel-wise depth from MVS point cloud to regulate the sparse view optimization. We directly adopt the results reported in the paper.

\subsubsection{Sat-Mesh}
Sat-Mesh\cite{qu_sat-mesh_2023} improves geometric reconstruction quality by leveraging multi-view texture consistency at the same locations in images. As the source code for this method is not publicly available, we use the data reported by the authors in the paper for comparison. In addition, this method is not included in the qualitative experiments for comparison.

\subsubsection{FVMD-ISRe and NeuS}
FVMD-ISRe\cite{zhang_fvmd-isre_2024} effectively reconstructs geometry by improving the positional encoding strategy, incorporating the sunlight direction input into the color MLP, and integrating an RPC-based ray-casting approach into NeuS\cite{wang2021neus} reconstruction. We use the same number of input images as our method, with a batch size set to 4096 and 100K iterations, while other settings follow the default configuration. The results of NeuS\cite{wang2021neus} follow the report on the FVMD-ISRe.

\subsection{Performance Comparison and Analysis}
\begin{table*}[!ht]
    \centering
    \caption{Quantitative experimental results of the reconstructed DSMs and Meshes in Jacksonville.}
    \setlength{\tabcolsep}{4pt} 
    \resizebox{1\textwidth}{!}{ 
    \begin{tabular}{l|ccc|ccc|ccc|ccc|ccc}
    \toprule
\multirow{2}{*}{\textbf{Method}} & \multicolumn{3}{c}{\textbf{JAX\_068}} & \multicolumn{3}{c}{\textbf{JAX\_175}} & \multicolumn{3}{c}{\textbf{JAX\_207}} & \multicolumn{3}{c}{\textbf{JAX\_214}} & \multicolumn{3}{c}{\textbf{JAX\_260}} \\
    \cmidrule(lr){2-4} \cmidrule(lr){5-7} \cmidrule(lr){8-10} \cmidrule(lr){11-13} \cmidrule(lr){14-16}
& MAE $\downarrow$ & MED $\downarrow$ & CD $\downarrow$ & MAE $\downarrow$ & MED $\downarrow$ & CD $\downarrow$ & MAE $\downarrow$ & MED $\downarrow$ & CD $\downarrow$ & MAE $\downarrow$ & MED $\downarrow$ & CD $\downarrow$ & MAE $\downarrow$ & MED $\downarrow$ & CD $\downarrow$ \\
    \midrule
S2P\cite{franchis2014pipeline}       &1.922            &0.944            &3.709             &\textbf{1.676}   &\textbf{0.536}   &\textbf{3.976}    &2.932            &\textbf{0.740}   &6.122             &2.887            &\underline{0.756}&4.463             &2.304            &\textbf{0.879}&1.676  \\
VisSat\cite{VisSat-2019}            &1.676            &1.013            &3.341             &2.644            &1.619            &4.594             &2.463            &1.280            &\underline{4.599} &2.364            &1.396            &3.675             &2.157            &1.240            &\underline{1.481} \\
S-NeRF\cite{kangle2021dsnerf}        &2.482            &1.831            &4.625             &3.876            &2.769            &6.633             &2.895            &2.055            &5.205             &5.460            &4.133            &7.458             &3.583            &2.916            &2.124 \\
Sat-NeRF\cite{mari2022sat}           &2.062            &1.330            &4.023             &3.274            &2.292            &5.791             &\textbf{2.128}   &1.163            &\textbf{4.324}    &4.600            &3.579            &6.551             &3.110            &2.411            &1.967 \\
SpS-NeRF\cite{zhang2023spsnerf}\textsuperscript{*} &-                &-                &-                 &-                &-                &-                 &-                &-                &-                 &2.860            &-                &-                 &-               &-                &-  \\
Sat-Mesh\cite{qu_sat-mesh_2023}\textsuperscript{*}      &1.146            &0.570            &-                 &-                &-                &-                 &-                &-                &-                 &2.022            &0.982            &-                 &-               &-                &-  \\
NeuS\cite{wang2021neus}\textsuperscript{*} &3.866            &-            &-                 &-                &-                &-                 &3.644                &-                &-                 &5.938            &-            &-                 &2.943               &-                &-  \\
FVMD-ISRe\cite{zhang_fvmd-isre_2024} &\underline{1.032}&\textbf{0.548}   &\underline{2.725} &2.132            &1.177            &4.413             &2.409            &\underline{1.017}&5.400             &\underline{1.980}&1.037            &\underline{3.516} &\underline{1.838}&1.052            &1.691  \\
Ours                                 &\textbf{1.030}   &\underline{0.549}&\textbf{2.654}    &\underline{1.833}&\underline{1.011}&\underline{4.020} &\underline{2.399}&1.214            &5.480             &\textbf{1.816}   &\textbf{0.732}   &\textbf{3.503}   &\textbf{1.620}   &\underline{0.961}&\textbf{1.333} \\
    \bottomrule
    \multicolumn{16}{l}{* The performance is derived from the authors’ report, with dashes indicating that no experimental results are available for the corresponding scenes.}\\
    \end{tabular}}
    \label{tab_2}
\end{table*}

\begin{table*}[!ht]
    \centering
    \caption{Quantitative experimental results of the reconstructed DSMs and Meshes in Omaha.}
    \setlength{\tabcolsep}{4pt} 
    \resizebox{1\textwidth}{!}{ 
    \begin{tabular}{l|ccc|ccc|ccc|ccc|ccc}
    \toprule
\multirow{2}{*}{\textbf{Method}} & \multicolumn{3}{c}{\textbf{OMA\_203}} & \multicolumn{3}{c}{\textbf{OMA\_212}} & \multicolumn{3}{c}{\textbf{OMA\_248}} & \multicolumn{3}{c}{\textbf{OMA\_287}} & \multicolumn{3}{c}{\textbf{OMA\_315}} \\
    \cmidrule(lr){2-4} \cmidrule(lr){5-7} \cmidrule(lr){8-10} \cmidrule(lr){11-13} \cmidrule(lr){14-16}
& MAE $\downarrow$ & MED $\downarrow$ & CD $\downarrow$ & MAE $\downarrow$ & MED $\downarrow$ & CD $\downarrow$ & MAE $\downarrow$ & MED $\downarrow$ & CD $\downarrow$ & MAE $\downarrow$ & MED $\downarrow$ & CD $\downarrow$ & MAE $\downarrow$ & MED $\downarrow$ & CD $\downarrow$ \\
    \midrule
S2P\cite{franchis2014pipeline}       &1.007            &\textbf{0.188}   &3.386              &1.541            &0.889            &3.550             &1.281            &0.553            &3.101             &\underline{0.964}&\textbf{0.310}   &\underline{2.829} &1.114            &\textbf{0.367}   &3.014 \\
VisSat\cite{VisSat-2019}            &1.023            &0.468            &2.895              &\textbf{0.753}   &\textbf{0.216}   &\underline{2.525} &1.154            &\textbf{0.538}   &2.865             &1.274            &0.639            &3.003             &\underline{1.043}&0.556            &\underline{2.889} \\
S-NeRF\cite{kangle2021dsnerf}        &2.771            &2.301            &5.333              &1.951            &1.319            &4.349             &4.279            &3.691            &7.494             &3.094            &2.633            &5.765             &2.205            &1.627            &4.425 \\
Sat-NeRF\cite{mari2022sat}           &3.011            &2.475            &5.628              &1.820            &1.321            &4.112             &3.435            &2.759            &6.512             &1.967            &1.472            &4.166             &1.959            &1.902            &4.087 \\
NeuS\cite{wang2021neus}\textsuperscript{*} &1.447            &-            &-              &1.314            &-            &-             &2.564            &-            &-             &17.594            &-            &-             &2.262            &-            &- \\
FVMD-ISRe\cite{zhang_fvmd-isre_2024} &\textbf{0.798}   &\underline{0.346}&\textbf{2.733}     &0.924            &\underline{0.478}&\textbf{2.508}    &\textbf{1.019}   &\underline{0.542}&\underline{2.727} &1.119            &0.589            &2.922             &1.072            &\underline{0.466}&2.920 \\
Ours                                 &\underline{0.930}&0.511            &\underline{2.870} &\underline{0.888}&0.529            &2.603             &\underline{1.037}&0.568            &\textbf{2.684}    &\textbf{0.894}   &\underline{0.405}&\textbf{2.624}    &\textbf{1.022}   &0.598            &\textbf{2.877} \\
    \bottomrule
    \multicolumn{16}{l}{* The performance is derived from the authors’ report, with dashes indicating that no experimental results are available for the corresponding scenes.}\\
    \end{tabular}}
    \label{tab_3}
\end{table*}

\begin{figure*}[!ht]
    \centering
    \includegraphics[width=1\linewidth]{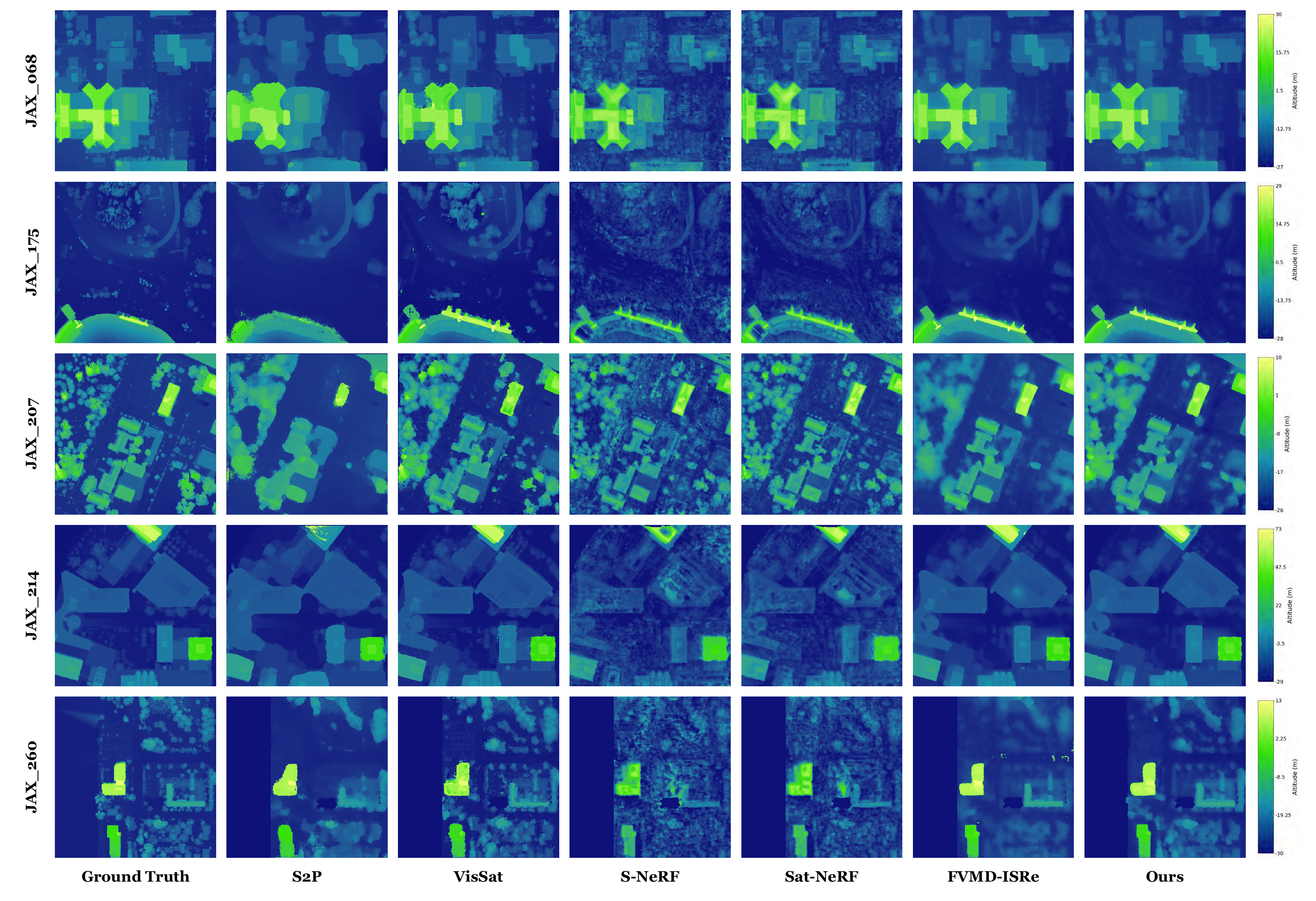}
    \caption{Qualitative experimental results of DSMs reconstructed from Jacksonville. Each row represents the DSMs obtained using different methods within the same scene, and each column corresponds to DSMs of different scenes reconstructed by the same method. The right side shows the altitude legend.}
    \label{fig:JAX_dsm}
\end{figure*}

\begin{figure*}[!ht]
    \centering
    \includegraphics[width=1\linewidth]{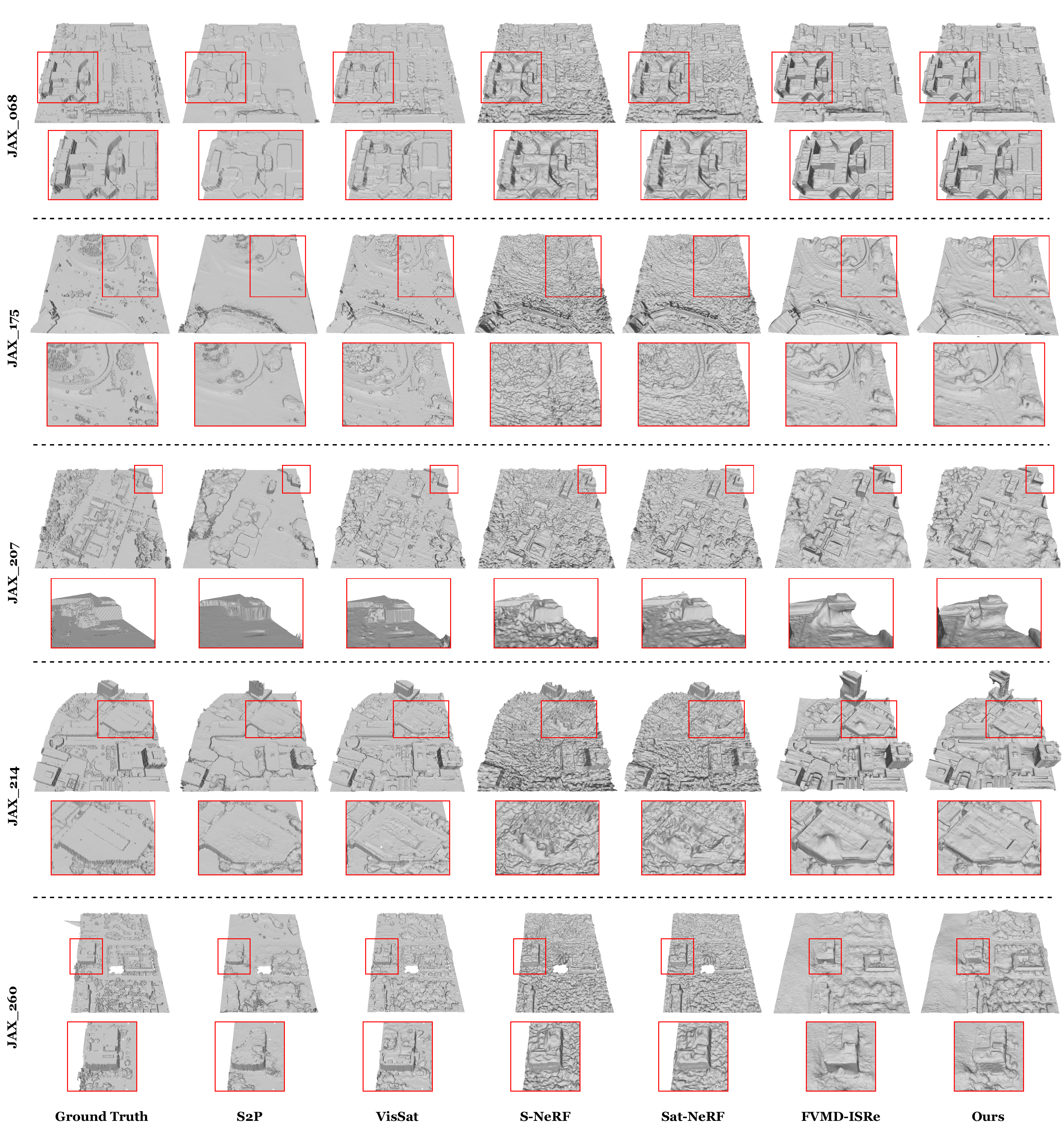}
    \caption{Qualitative experimental results of Mesh reconstructions in Jacksonville, with red boxes indicating zoomed-in areas. Except for FVMD-ISRe and our method using marching cubes to generate Meshes, other approaches generate Mesh models based on adjusted DSMs, resulting in missing sections due to referenced Ground truth.}
    \label{fig:JAX_mesh}
\end{figure*}

\begin{figure*}[!ht]
    \centering
    \includegraphics[width=1\linewidth]{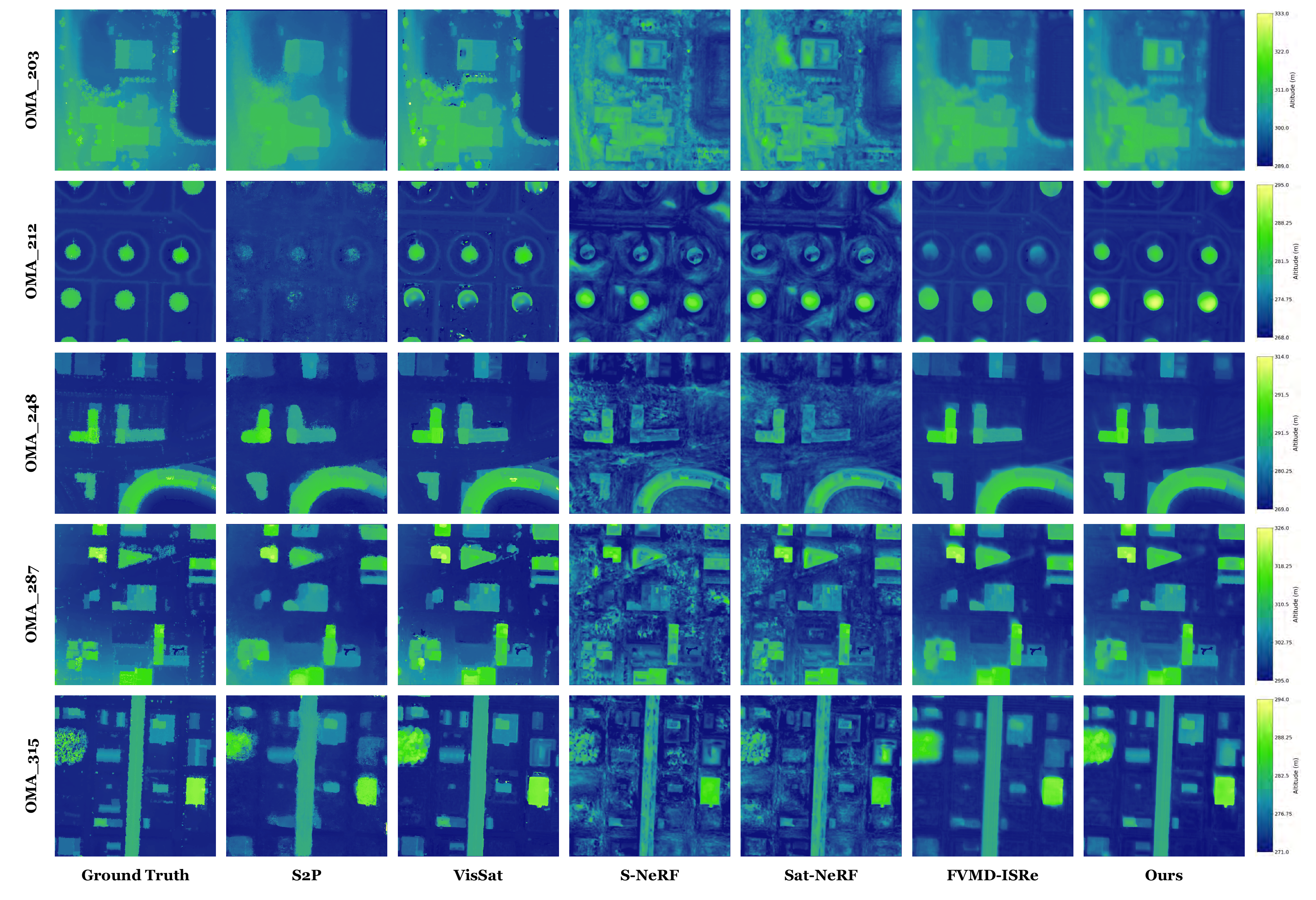}
    \caption{Qualitative experimental results of DSMs reconstructed from Omaha. Each row represents the DSMs obtained using different methods within the same scene, and each column corresponds to DSMs of different scenes reconstructed by the same method. The right side shows the altitude legend.}
    \label{fig:OMA_dsm}
\end{figure*}

\begin{figure*}[!ht]
    \centering
    \includegraphics[width=1\linewidth]{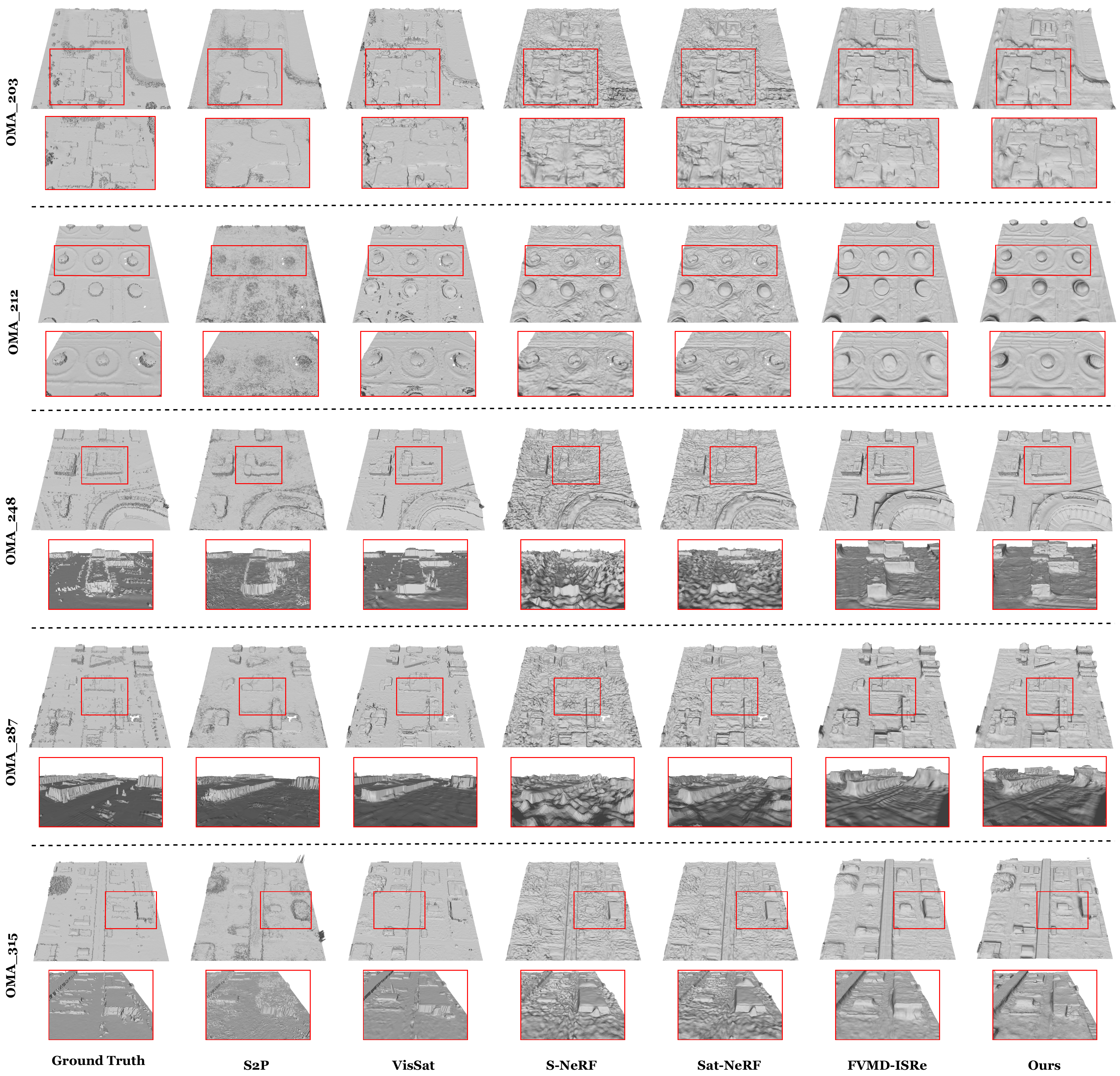}
    \caption{Qualitative experimental results of Mesh reconstructions in Omaha, with red boxes indicating zoomed-in areas. Except for FVMD-ISRe and our method using marching cubes to generate Meshes, other approaches generate Mesh models based on adjusted DSMs, resulting in missing sections due to referenced Ground truth.}
    \label{fig:OMA_mesh}
\end{figure*}

Our quantitative experiments on the Jacksonville subset of the DFC2019 dataset are shown in Table. \ref{tab_2}, demonstrating that our method achieves the best results across most scenes. Qualitative results are illustrated in Fig. \ref{fig:JAX_dsm} and Fig. \ref{fig:JAX_mesh}, where Fig. \ref{fig:JAX_dsm} shows the DSMs reconstructed by different methods, and Fig. \ref{fig:JAX_mesh} presents the reconstructed Mesh models. Our method achieves consistently favourable results across all scenes.

In Table. \ref{tab_2}, bold values in the table indicate the best results and the underlined values represent the second-best results. Arrows indicate the direction in which the evaluation metrics favor better results. In the JAX\_068 and JAX\_214 scenes, our method achieved the best MAE and CD results and was close to the best method in MED. In the JAX\_175 scene, due to mismatches between the ground truth DSM and the real scene in the image, our method still produced the second-best results across all evaluation metrics, indicating that our method performs well in most regions of the scene with only slight discrepancies from the ground truth. In the JAX\_207 and JAX\_260 scenes, large vegetation and water surfaces affected the depth model’s depth estimation, leading to increased depth estimation errors that negatively impacted the reconstruction. In the image, the surface of the water will appear local high reflection; Vegetation has no regular geometric structure, and its colors are relatively homogeneous and messy. This is not conducive to the depth model to estimate a reasonable relative depth. However, our method still achieved relatively good results across all evaluation metrics.

Based on the reconstructed DSMs in Jacksonville shown in Fig. \ref{fig:JAX_dsm}. For example in JAX\_068, S2P exhibits extensive connections in its reconstruction, failing to reflect the actual appearance of buildings accurately. The building shapes are incorrect, and the edges have reconstruction errors. VisSat uses point clouds generated from structure-from-motion and projects them onto the ground to produce DSMs. Since the point clouds are not generated through dense matching, some pixels lack point cloud values during projection, resulting in holes in the DSM. Additionally, the point clouds contain some points with significant errors, leading to height discontinuities in the DSM after projection, such as the abrupt height change seen in the figure's upper left corner of the star-shaped building. In S-NeRF and Sat-NeRF, the reconstructed DSMs exhibit height fluctuations across the entire scene. This occurs because, although both S-NeRF and Sat-NeRF decompose the scene’s lighting, the DSM height calculation relies on volumetric density used for new view synthesis, which does not strictly align with the geometric surface. Sat-NeRF, as an extension of S-NeRF, outperforms S-NeRF in all evaluation metrics. Both FVMD-ISRe and our method produced relatively high-quality DSM results, preserving the correct geometric features. Notably, in the JAX\_214 scene, other neural radiance field-based methods exhibited unreasonable height issues on the top surface of the parking lot within the red box during reconstruction. Our method, however, ensures geometric accuracy over large planar regions due to the incorporation of depth and normal constraints. In other scenes, the aforementioned issues with the comparison methods persist. Both S2P and VisSat fail to produce reasonable geometric distributions. The depth estimated by the S-NeRF series methods remains unstable. Although the DSM calculated by FVMD-ISRe is similar to our method’s, it requires significant computational time and memory, which imposes certain hardware requirements.

Comparing the Meshes generated by various methods in Fig. \ref{fig:JAX_mesh}, our method produces the most refined mesh, recovering intricate surface details of the scene while ensuring geometric accuracy. Since S2P and VisSat calculate point clouds through geometric methods to generate Meshes, noticeable holes and erroneous points appear in their Mesh results. Explicit geometry-based calculation methods strictly adhere to mathematical principles, but building occlusions and lighting effects can cause discrepancies between the pixel’s spatial position and its image pixel values. This results in spatial positions deviating from their true locations, as seen in JAX\_068, where the left side of the red-boxed building’s podium exhibits height estimation errors. For S-NeRF and Sat-NeRF, due to the inherent misalignment of volumetric density with the actual surface, the generated Mesh appears chaotic across the entire scene, failing to produce accurate geometric structures. Sat-NeRF, after decoupling direct lighting from ambient lighting, shows some improvement over S-NeRF in scene reconstruction.

For FVMD-ISRe and our method, we enhanced surface reconstruction by incorporating depth and normal constraints, along with progressive training, ensuring geometric accuracy while optimizing planar distributions. In JAX\_068, our method reconstructed a more realistic surface, achieving smoother plane distributions while preserving subtle texture variations, effectively capturing corresponding surface undulations. In JAX\_175, the road surface reconstructed by FVMD-ISRe appeared blurry due to unstable normal orientations, whereas our method delivered sharp results, reflecting the actual smoothness of the road surface. In JAX\_207, the building indicated by the red box had recessed regions in the reconstruction due to limited visibility of the building’s facades in the images. Our method, leveraging progressive training and explicit geometric constraints, mitigated such recesses or hollow areas in the planes. In JAX\_214, the depth constraint significantly improved scene geometry, allowing our method to accurately reconstruct the rooftop of the building in the red box. This ensured smooth planar distributions and corrected surface estimation errors introduced by relying solely on RGB information. In JAX\_260, the building’s side facade reconstructed by FVMD-ISRe exhibited erroneous connections between the building surface and nearby vegetation on the ground. Our method, supported by depth priors, reasonably inferred the connection between the building and the ground, reducing the influence of surrounding vegetation on reconstruction quality.

The quantitative experimental results for Omaha on the DFC2019 dataset are shown in Table. \ref{tab_3}, where it can be seen that our method achieved the best performance in most of the scenes. The qualitative experiments are shown in Fig. \ref{fig:OMA_dsm} and Fig. \ref{fig:OMA_mesh}, where Fig. \ref{fig:OMA_dsm} displays the DSMs reconstructed by different approaches, and Fig. \ref{fig:OMA_mesh} shows the reconstructed Mesh. Our method evidently yields the best DSM and Mesh reconstruction results.

In the OMA\_203 and OMA\_212 scenes, the altitude range between the lowest and the highest is minimal, and the texture information is sparse. This leads to a reduced number of spatial points obtained through spatial triangulation. The decrease in point count may result in increased errors in depth fusion, which negatively impacts the reconstruction quality. Additionally, overly bright reflective light in the images causes depth estimation errors in the depth model, further degrading the reconstruction quality. On the other hand, in scenes like OMA\_248, OMA\_287, and OMA\_315, which contain more man-made structures like buildings, our method achieves the best results across all evaluation metrics.

As shown in Fig. \ref{fig:OMA_dsm}, in the reconstruction of the OMA\_203 scene, although explicit geometry-based methods show significant height variations and missing values in the DSM, the large-scale flat regions reconstructed are relatively reasonable. The S-NeRF method still exhibits its inherent height fluctuation problem. There is little difference between the DSM reconstructions of FVMD-ISRe and our method, both maintaining spatial continuity. In OMA\_212, all other methods failed to reconstruct some of the oil storage tanks due to the strong reflective light interference, while our method, although experiencing height displacement in the lower part of the tanks, successfully reconstructed all tank structures. In the building scenes of OMA\_248, OMA\_287, and OMA\_315, our method achieved the results closest to the ground truth across all evaluations.

As shown in Fig. \ref{fig:OMA_mesh}, the Mesh reconstructed in the Omaha scenes highlights the effectiveness of our method. In OMA\_203, our approach successfully reconstructed the most accurate rooftop in the red-boxed area, capturing subtle height variations visible in the ground truth. Compared to FVMD-ISRe, our method exhibited more consistent normal orientations, resulting in smoother surfaces for planes such as grass fields and playgrounds. In OMA\_212, the reconstructed storage tanks in the central region of the Mesh demonstrate the superiority of our method. Our method accurately preserved the geometric details of the tanks under correct geometry constraints, while other methods exhibited poor reconstruction quality or outright failure. This underscores the guiding role of depth constraints in improving geometric reconstruction accuracy. In OMA\_248 and OMA\_287, the sharp transitions between planar surfaces in the geometric structures reconstructed by our method were notably more distinct than those of FVMD-ISRe, owing to the integration of depth and normal guidance. This allowed for better planar representation in the reconstructed buildings while retaining rich geometric details. Finally, in OMA\_315, our method outperformed all compared methods in localized reconstruction quality, further demonstrating its robustness and effectiveness.

\begin{table*}[!ht]
    \centering
    \setlength{\tabcolsep}{5pt} 
    \caption{Training time and GPU memory usage statistics for the JAX scenes}
    \begin{tabular}{l|l|cccccccccc|cc}
    \toprule
\multicolumn{2}{c}{\textbf{Method}} & \multicolumn{2}{c}{\textbf{JAX\_068}} & \multicolumn{2}{c}{\textbf{JAX\_175}} & \multicolumn{2}{c}{\textbf{JAX\_207}} & \multicolumn{2}{c}{\textbf{JAX\_214}} & \multicolumn{2}{c}{\textbf{JAX\_260}} & \multicolumn{2}{c}{\textbf{Mean}}\\
    \midrule
Stereo Matching & S2P\cite{franchis2014pipeline}      & $\approx$0.4  &-      & $\approx$0.2  &-      & $\approx$0.3  &-      & $\approx$0.5  &-      & $\approx$0.2  &-      & 0.32  &- \\
    \midrule
\multirow{3}{*}{Neural Recon}&Sat-NeRF\cite{mari2022sat}          & $\approx$8.0  & 14822 & $\approx$7.1  & 15966 & $\approx$6.7  & 16942 & $\approx$6.8  & 14822 & $\approx$6.9  & 14822 & 7.1   & 15475 \\
&FVMD-ISRe\cite{zhang_fvmd-isre_2024}& $\approx$10.2 & 22708 & $\approx$10.1 & 22482 & $\approx$10.4 & 22482 & $\approx$10.3 & 22462 & $\approx$10.1 & 22614 & 10.22 & 22550\\
&Ours                                & $\approx$\textbf{5.6}  & \textbf{12356} & $\approx$\textbf{5.7}  & \textbf{12858} & $\approx$\textbf{5.4}  & \textbf{12860} & $\approx$\textbf{5.7}  & \textbf{12840} & $\approx$\textbf{5.6}  & \textbf{12876} & \textbf{5.6} & \textbf{12758}\\
    \bottomrule
    \multicolumn{13}{l}{The time statistics are measured in hours, and the GPU memory usage is measured in MB.} \\
    \end{tabular}
    \label{tab_4}
\end{table*}

The quantitative comparison of training time and memory consumption between our method and others is shown in Table. \ref{tab_4}. The stereo matching method S2P, which does not utilize GPU resources, is excluded from memory usage statistics. However, it is the fastest among all compared methods in terms of runtime. Nevertheless, its DSM accuracy is limited, and it cannot directly produce a Mesh. Among neural reconstruction methods, our approach achieves the lowest training time and memory consumption, followed by Sat-NeRF. FVMD-ISRe exhibits the highest resource consumption. Notably, Sat-NeRF cannot directly generate Meshes. Furthermore, our method surpasses others in reconstruction quality, demonstrating that it excels in both reconstruction efficiency and quality compared to the other methods.

\subsection{Ablation Studies}
\begin{figure*}[!ht]
    \centering
    \includegraphics[width=1\linewidth]{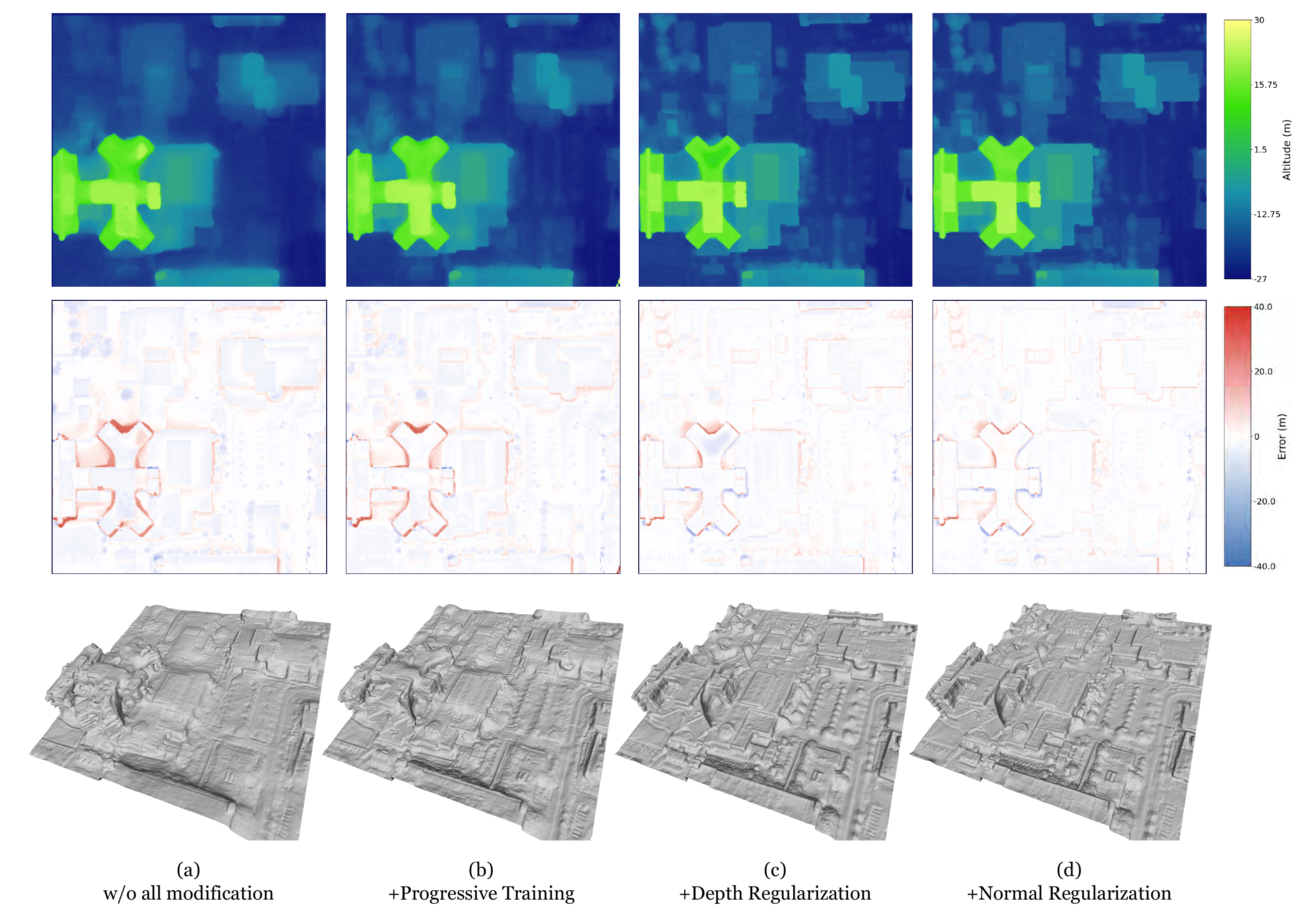}
    \caption{Comparison of qualitative results in the ablation experiments. Column (a) shows the reconstruction results using only the hash grid; columns (b), (c), and (d) sequentially represent the reconstruction results after progressively adding progressive training, depth regularization, and normal consistency regularization. The first row shows the reconstructed DSMs, the middle row shows DSM errors, and the last row shows the reconstructed Mesh.}
    \label{fig:ablation}
\end{figure*}

In this section, we validate the impact of the proposed depth regularization (DepthReg), normal regularization based on surface normal consistency (NormalReg), and progressive training (PTrain) on the reconstruction quality. We use the JAX\_068 scene as the dataset for the ablation study, and the results of the ablation experiments are shown in Fig. \ref{fig:ablation} and Table. \ref{tab_5}.

\begin{table}[!htbp]
    \centering
    \caption{Quantitative comparison results of the ablation experiments on JAX\_068}
    \begin{tabular}{ccc|ccc}
    \toprule
    \textbf{PTrain} & \textbf{DepthReg} & \textbf{NormalReg} & MAE $\downarrow$ & MED $\downarrow$ & CD $\downarrow$ \\
    \midrule
    && & 1.790 & 1.078 & 4.694 \\
    $\checkmark$ & & & 1.599 & 1.009 & 3.578 \\
    $\checkmark$ & $\checkmark$ & & 1.127 & 0.639 & 2.962 \\
    $\checkmark$ & $\checkmark$ & $\checkmark$ & \textbf{1.030} & \textbf{0.549} & \textbf{2.654} \\
    \bottomrule
    \end{tabular}
    \label{tab_5}
\end{table}

\subsubsection{Progressive Training}
As shown in Fig. \ref{fig:ablation}(a), using only the hash grid for reconstructing ground models results in many voids on the side surfaces of buildings, with overly smooth transitions between building roofs, side surfaces, and the ground, leading to significant discrepancies between the reconstruction and the real appearance of buildings. After incorporating the progressive training strategy, the model is guided to prioritize learning the low-frequency coarse geometric information of the scene before refining high-frequency local details, enabling heuristic model training. Fig. \ref{fig:ablation}(b) and Table. \ref{tab_5} show that the reconstruction quality improves significantly.

\subsubsection{Depth Regularization}
\begin{figure}[!ht]
    \centering
    \includegraphics[width=0.9\linewidth]{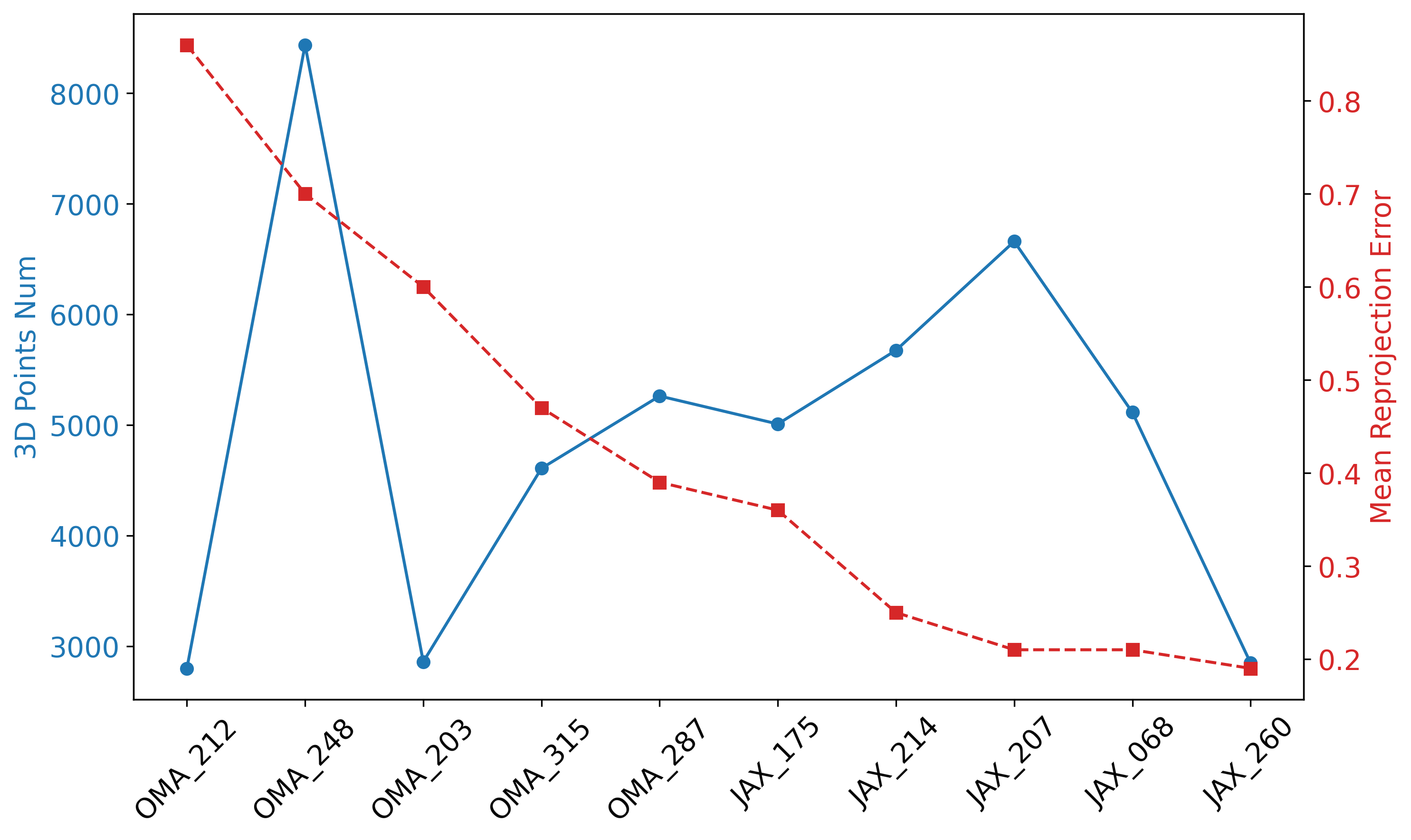}
    \caption{The number of spatial feature points and the average reprojection error for each scene obtained after bundle adjustment. The average projection error of spatial sparse points is negatively correlated with the performance of our method.}
    \label{fig:ablation_depth}
\end{figure}

\begin{figure}[!ht]
    \centering
    \includegraphics[width=0.9\linewidth]{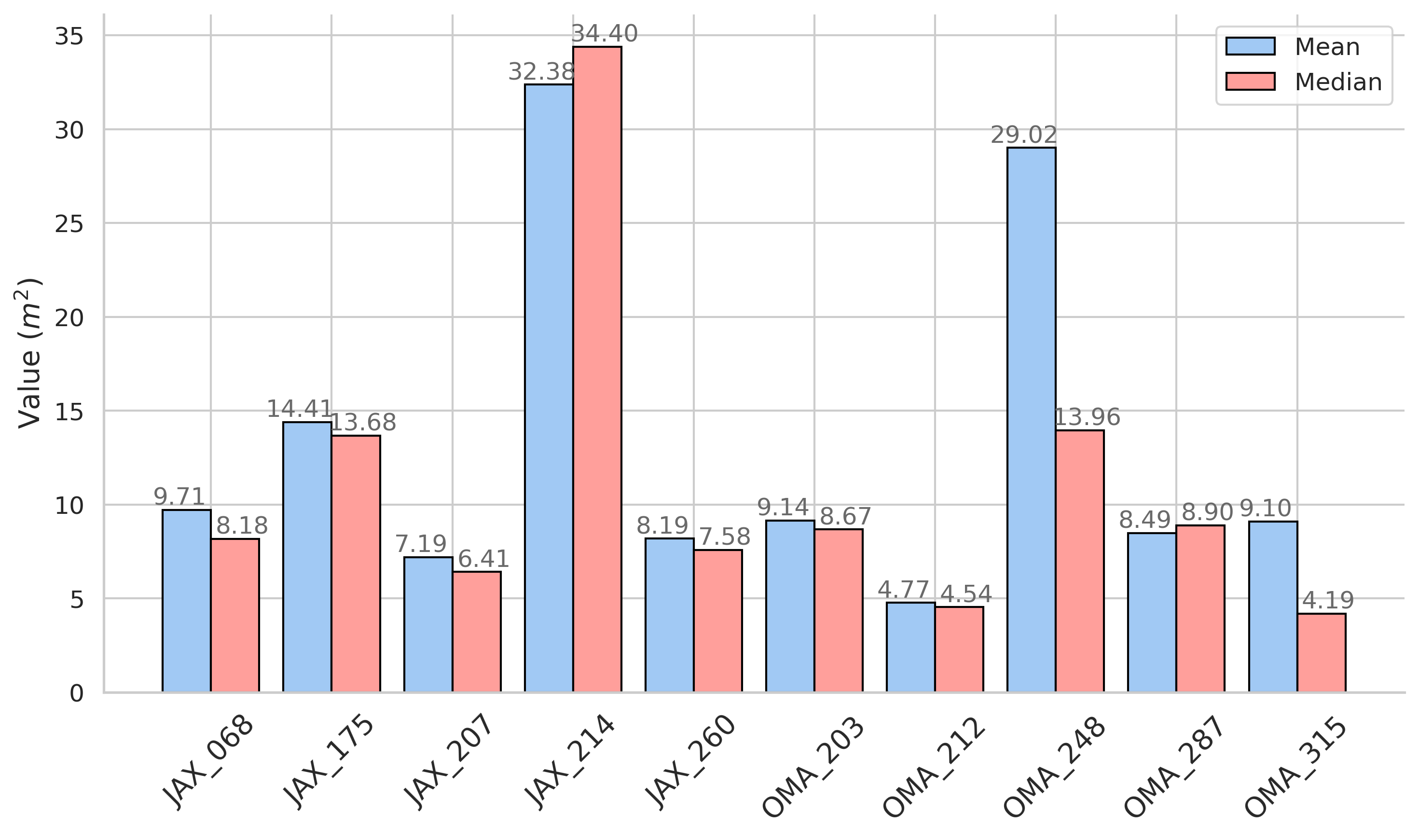}
    \caption{Least square depth fusion errors of each scene in DFC2019 dataset.}
    \label{fig:ablation_depth_fusion}
\end{figure}

To address issues such as smooth connections between building planes and poor reconstruction quality in the model, depth constraints were introduced to explicitly supervise the opacity distribution along the rays, thereby producing planes with accurate heights and sharper plane transitions. As shown in Fig. \ref{fig:ablation}(c) and Table. \ref{tab_5}, adding depth constraints significantly improved both qualitative and quantitative reconstruction evaluations.

Notably, due to imbalances in lighting within the scene, errors in surface position estimation occurred when reconstructing building rooftops, necessitating the subsequent introduction of normal constraints. Fig. \ref{fig:ablation_depth} lists the number of retained spatial 3D points and their reprojection errors in 2D images after bundle adjustment for each scene. It can be observed that the number of spatial points and their reprojection errors impact the reconstruction results of our method. The number of 3D points positively correlates with reconstruction quality, while reprojection error is negatively correlated. In addition, we recorded the final least square losses means and medians of the depth fusion in Fig. \ref{fig:ablation_depth_fusion}. In scenarios with large height differences or complicated geometry, such as JAX\_214 and OMA\_248, although the depth estimation model can obtain the relative elevation, this relative elevation cannot accurately correspond to the real height difference. However, in the vast majority of scenarios, the error of depth fusion is relatively small and can guide the model to converge to the correct geometry, thereby alleviating the ambiguity of optimization brought about by fluctuations in pixel photometric values.

The Depth Anything V2\cite{depth_anything_v2} model we use is one of the SOTA models in the computer vision community, which is trained by real-world images. Despite the limitations of using computer vision models in satellite images, the experiments presented above prove that the computer vision model can achieve a reasonable relative depth estimation in satellite images. Since this study focuses on surface reconstruction, the depth estimation content is not ablated.

\subsubsection{Normal Regularization}
Due to significant lighting variations, depth guidance's effectiveness in estimating accurate surfaces is limited. By introducing surface normal consistency regularization, the quality of planar reconstruction in areas with extreme lighting was further improved. Moreover, compared with models reconstructed by other methods, our proposed normal regularization smooths the distribution of normals on planes, such as building rooftops or ground surfaces, without compromising scene details, ensuring their orientations remain consistent within the same plane. As shown in Fig. \ref{fig:ablation}(d), our complete method achieves the best reconstruction result.

\section{Discussion}
\subsection{Comparison between Nerf-based and Traditional Methods}
Traditional methods for processing satellite imagery primarily rely on sparse or stereo matching techniques. However, these approaches depend on complete visual correspondences and highly accurate matching, which often result in discontinuities and voids in the reconstructed point clouds and meshes. Such incomplete and fragmented outputs are typically unsuitable for downstream applications, especially in scenarios that demand high-fidelity 3D representations. Moreover, traditional methods are prone to the influence of dynamic objects, often introducing outliers into the reconstruction.

In contrast, NeRF-based methods offer a significant advantage by generating watertight and continuous meshes that are structurally coherent and largely free from such artifacts. These meshes can be directly manipulated using standard 3D modeling software, thereby enhancing their practical applicability.

Although NeRF-based approaches generally require longer processing times, the improved quality and completeness of the reconstructed geometry often justify the additional computational cost. With ongoing advancements in hardware, this gap is expected to gradually diminish.

\section{Conclusion}\label{conclusion}
In this paper, we propose Sat-DN, a method capable of reconstructing DSM and Mesh from multi-view, multi-temporal satellite imagery. To address inherent challenges in satellite imagery, such as poor visibility of building facades, variations in lighting and color between images, and the prevalence of low-texture regions, we build upon previous research to introduce a multi-resolution hash grid reconstruction framework with a progressive training strategy. Additionally, we incorporate real-scale depth guidance and surface normal consistency constraints. Extensive experiments on the DFC2019 dataset demonstrate the superiority of our method over others.

Although our method is relatively robust to lighting, there are still errors when estimating surfaces with overexposed areas. In the future, we will focus on decoupling lighting in surface reconstruction and further investigate neural reconstruction methods for vegetation and water surface areas.

\section*{Acknowledgments}
We would like to thank the researchers from IEEE GRSS and Johns Hopkins University for providing the DFC2019 dataset, the researchers from the University of Hong Kong and Tiktok for providing the Depth Anything V2 model, and Yuan Zhou, Shenhong Li, Yang Lei, and Yixuan Liu for their discussion and support.

\bibliographystyle{IEEEtran}
\bibliography{ref}

\begin{thebibliography}{10}
\providecommand{\url}[1]{#1}
\csname url@samestyle\endcsname
\providecommand{\newblock}{\relax}
\providecommand{\bibinfo}[2]{#2}
\providecommand{\BIBentrySTDinterwordspacing}{\spaceskip=0pt\relax}
\providecommand{\BIBentryALTinterwordstretchfactor}{4}
\providecommand{\BIBentryALTinterwordspacing}{\spaceskip=\fontdimen2\font plus
\BIBentryALTinterwordstretchfactor\fontdimen3\font minus \fontdimen4\font\relax}
\providecommand{\BIBforeignlanguage}[2]{{%
\expandafter\ifx\csname l@#1\endcsname\relax
\typeout{** WARNING: IEEEtran.bst: No hyphenation pattern has been}%
\typeout{** loaded for the language `#1'. Using the pattern for}%
\typeout{** the default language instead.}%
\else
\language=\csname l@#1\endcsname
\fi
#2}}
\providecommand{\BIBdecl}{\relax}
\BIBdecl

\bibitem{zhao2023review}
L.~Zhao, H.~Wang, Y.~Zhu, and M.~Song, ``A review of 3d reconstruction from high-resolution urban satellite images,'' \emph{International Journal of Remote Sensing}, vol.~44, no.~2, pp. 713--748, 2023.

\bibitem{facciolo2017automatic}
G.~Facciolo, C.~De~Franchis, and E.~Meinhardt-Llopis, ``Automatic 3d reconstruction from multi-date satellite images,'' in \emph{2017 IEEE Conference on Computer Vision and Pattern Recognition Workshops (CVPRW)}, 2017, pp. 1542--1551.

\bibitem{Leotta_2019_CVPR_Workshops}
M.~J. Leotta, C.~Long, B.~Jacquet, M.~Zins, D.~Lipsa, J.~Shan, B.~Xu, Z.~Li, X.~Zhang, S.-F. Chang, M.~Purri, J.~Xue, and K.~Dana, ``Urban semantic 3d reconstruction from multiview satellite imagery,'' in \emph{Proceedings of the IEEE/CVF Conference on Computer Vision and Pattern Recognition (CVPR) Workshops}, June 2019.

\bibitem{zhang_leveraging_2019}
K.~Zhang, N.~Snavely, and J.~Sun, ``\BIBforeignlanguage{en}{Leveraging {Vision} {Reconstruction} {Pipelines} for {Satellite} {Imagery}},'' in \emph{\BIBforeignlanguage{en}{2019 {IEEE}/{CVF} {International} {Conference} on {Computer} {Vision} {Workshop} ({ICCVW})}}.\hskip 1em plus 0.5em minus 0.4em\relax IEEE, Oct. 2019, pp. 2139--2148.

\bibitem{mari2022sat}
R.~Mar{\'\i}, G.~Facciolo, and T.~Ehret, ``{Sat-NeRF}: Learning multi-view satellite photogrammetry with transient objects and shadow modeling using {RPC} cameras,'' in \emph{2022 IEEE/CVF Conference on Computer Vision and Pattern Recognition Workshops (CVPRW)}, 2022, pp. 1310--1320.

\bibitem{mildenhall2021nerf}
B.~Mildenhall, P.~P. Srinivasan, M.~Tancik, J.~T. Barron, R.~Ramamoorthi, and R.~Ng, ``Nerf: Representing scenes as neural radiance fields for view synthesis,'' \emph{Communications of the ACM}, vol.~65, no.~1, pp. 99--106, 2021.

\bibitem{barron2021mipnerf}
J.~T. Barron, B.~Mildenhall, M.~Tancik, P.~Hedman, R.~Martin-Brualla, and P.~P. Srinivasan, ``Mip-nerf: A multiscale representation for anti-aliasing neural radiance fields,'' \emph{ICCV}, 2021.

\bibitem{derksen2021shadow}
D.~Derksen and D.~Izzo, ``Shadow neural radiance fields for multi-view satellite photogrammetry,'' in \emph{Proceedings of the IEEE/CVF Conference on Computer Vision and Pattern Recognition}, 2021, pp. 1152--1161.

\bibitem{Mari_2023_eo}
R.~Mar{\'\i}, G.~Facciolo, and T.~Ehret, ``Multi-date earth observation nerf: The detail is in the shadows,'' in \emph{Proceedings of the IEEE/CVF Conference on Computer Vision and Pattern Recognition (CVPR) Workshops}, June 2023, pp. 2034--2044.

\bibitem{zhang_fvmd-isre_2024}
C.~Zhang, Y.~Yan, C.~Zhao, N.~Su, and W.~Zhou, ``Fvmd-isre: 3-d reconstruction from few-view multidate satellite images based on the implicit surface representation of neural radiance fields,'' \emph{IEEE Transactions on Geoscience and Remote Sensing}, vol.~62, pp. 1--14, 2024.

\bibitem{schoenberger2016sfm}
J.~L. Sch\"{o}nberger and J.-M. Frahm, ``Structure-from-motion revisited,'' in \emph{Conference on Computer Vision and Pattern Recognition (CVPR)}, 2016.

\bibitem{wang2021neus}
P.~Wang, L.~Liu, Y.~Liu, C.~Theobalt, T.~Komura, and W.~Wang, ``Neus: Learning neural implicit surfaces by volume rendering for multi-view reconstruction,'' \emph{arXiv preprint arXiv:2106.10689}, 2021.

\bibitem{martin2021nerfw}
R.~Martin-Brualla, N.~Radwan, M.~S. Sajjadi, J.~T. Barron, A.~Dosovitskiy, and D.~Duckworth, ``Nerf in the wild: Neural radiance fields for unconstrained photo collections,'' in \emph{Proceedings of the IEEE/CVF conference on computer vision and pattern recognition}, 2021, pp. 7210--7219.

\bibitem{rematas2022urban}
K.~Rematas, A.~Liu, P.~P. Srinivasan, J.~T. Barron, A.~Tagliasacchi, T.~Funkhouser, and V.~Ferrari, ``Urban radiance fields,'' in \emph{Proceedings of the IEEE/CVF Conference on Computer Vision and Pattern Recognition}, 2022, pp. 12\,932--12\,942.

\bibitem{Sabour_2023_robustnerf}
S.~Sabour, S.~Vora, D.~Duckworth, I.~Krasin, D.~J. Fleet, and A.~Tagliasacchi, ``Robustnerf: Ignoring distractors with robust losses,'' in \emph{Proceedings of the IEEE/CVF Conference on Computer Vision and Pattern Recognition (CVPR)}, June 2023, pp. 20\,626--20\,636.

\bibitem{Fu2022GeoNeus}
Q.~Fu, Q.~Xu, Y.-S. Ong, and W.~Tao, ``Geo-neus: Geometry-consistent neural implicit surfaces learning for multi-view reconstruction,'' \emph{Advances in Neural Information Processing Systems (NeurIPS)}, 2022.

\bibitem{Wang2022NeuS2}
Y.~Wang, Q.~Han, M.~Habermann, K.~Daniilidis, C.~Theobalt, and L.~Liu, ``Neus2: Fast learning of neural implicit surfaces for multi-view reconstruction,'' \emph{2023 IEEE/CVF International Conference on Computer Vision (ICCV)}, pp. 3272--3283, 2022.

\bibitem{kangle2021dsnerf}
K.~Deng, A.~Liu, J.-Y. Zhu, and D.~Ramanan, ``Depth-supervised {NeRF}: Fewer views and faster training for free,'' in \emph{Proceedings of the IEEE/CVF Conference on Computer Vision and Pattern Recognition (CVPR)}, June 2022.

\bibitem{guo2023streetsurf}
J.~Guo, N.~Deng, X.~Li, Y.~Bai, B.~Shi, C.~Wang, C.~Ding, D.~Wang, and Y.~Li, ``Streetsurf: Extending multi-view implicit surface reconstruction to street views,'' \emph{arXiv preprint arXiv:2306.04988}, 2023.

\bibitem{depth_anything_v2}
L.~Yang, B.~Kang, Z.~Huang, Z.~Zhao, X.~Xu, J.~Feng, and H.~Zhao, ``Depth anything v2,'' \emph{arXiv:2406.09414}, 2024.

\bibitem{mueller2022instant}
T.~M\"uller, A.~Evans, C.~Schied, and A.~Keller, ``Instant neural graphics primitives with a multiresolution hash encoding,'' \emph{ACM Trans. Graph.}, vol.~41, no.~4, pp. 102:1--102:15, Jul. 2022.

\bibitem{dfc2019data}
B.~Le~Saux, N.~Yokoya, R.~Hänsch, and M.~Brown, ``Data fusion contest 2019 (dfc2019),'' 2019.

\bibitem{duan2016towards}
L.~Duan and F.~Lafarge, ``Towards large-scale city reconstruction from satellites,'' in \emph{ECCV 2016: 14th European Conference, Amsterdam, The Netherlands, October 11-14, 2016, Proceedings, Part V 14}.\hskip 1em plus 0.5em minus 0.4em\relax Springer, 2016, pp. 89--104.

\bibitem{zhao2021double}
L.~Zhao, Y.~Liu, C.~Men, and Y.~Men, ``Double propagation stereo matching for urban 3-d reconstruction from satellite imagery,'' \emph{IEEE Transactions on Geoscience and Remote Sensing}, vol.~60, pp. 1--17, 2021.

\bibitem{stucker2022resdepth}
C.~Stucker and K.~Schindler, ``Resdepth: A deep residual prior for 3d reconstruction from high-resolution satellite images,'' \emph{ISPRS Journal of Photogrammetry and Remote Sensing}, vol. 183, pp. 560--580, 2022.

\bibitem{gao2023general}
J.~Gao, J.~Liu, and S.~Ji, ``A general deep learning based framework for 3d reconstruction from multi-view stereo satellite images,'' \emph{ISPRS Journal of Photogrammetry and Remote Sensing}, vol. 195, pp. 446--461, 2023.

\bibitem{OROZCO2016121}
R.~Orozco, C.~Loscos, I.~Martin, and A.~Artusi, ``Chapter 4 - multiview hdr video sequence generation,'' in \emph{High Dynamic Range Video}, F.~Dufaux, P.~{Le Callet}, R.~K. Mantiuk, and M.~Mrak, Eds.\hskip 1em plus 0.5em minus 0.4em\relax Academic Press, 2016, pp. 121--138.

\bibitem{kolmogorov2001computing}
V.~Kolmogorov and R.~Zabih, ``Computing visual correspondence with occlusions using graph cuts,'' in \emph{Proceedings Eighth IEEE International Conference on Computer Vision. ICCV 2001}, vol.~2.\hskip 1em plus 0.5em minus 0.4em\relax IEEE, 2001, pp. 508--515.

\bibitem{hirschmuller2005accurate}
H.~Hirschmuller, ``Accurate and efficient stereo processing by semi-global matching and mutual information,'' in \emph{2005 IEEE computer society conference on computer vision and pattern recognition (CVPR'05)}, vol.~2.\hskip 1em plus 0.5em minus 0.4em\relax IEEE, 2005, pp. 807--814.

\bibitem{hirschmuller2007stereo}
{H. Hirschmuller}, ``Stereo processing by semiglobal matching and mutual information,'' \emph{IEEE Transactions on pattern analysis and machine intelligence}, vol.~30, no.~2, pp. 328--341, 2007.

\bibitem{Facciolo2015MGMAS}
G.~Facciolo, C.~de~Franchis, and E.~Meinhardt, ``Mgm: A significantly more global matching for stereovision,'' in \emph{British Machine Vision Conference}, 2015.

\bibitem{lastilla2019}
L.~Lastilla, R.~Ravanelli, F.~Fratarcangeli, M.~Di~Rita, A.~Nascetti, and M.~Crespi, ``Foss4g date for dsm generation: Sensitivity analysis of the semi-global block matching parameters,'' \emph{ISPRS - International Archives of the Photogrammetry, Remote Sensing and Spatial Information Sciences}, vol. XLII-2/W13, pp. 67--72, 06 2019.

\bibitem{ghuffar2016satellite}
S.~Ghuffar, ``Satellite stereo based digital surface model generation using semi global matching in object and image space,'' \emph{ISPRS Annals of the Photogrammetry, Remote Sensing and Spatial Information Sciences}, vol.~3, pp. 63--68, 2016.

\bibitem{dumas2022improving}
L.~Dumas, V.~Defonte, Y.~Steux, and E.~Sarrazin, ``Improving pairwise dsm with 3sgm: A semantic segmentation for sgm using an automatically refined neural network,'' \emph{ISPRS Annals of the Photogrammetry, Remote Sensing and Spatial Information Sciences}, vol.~2, pp. 167--175, 2022.

\bibitem{yang2020novel}
W.~Yang, X.~Li, B.~Yang, and Y.~Fu, ``A novel stereo matching algorithm for digital surface model (dsm) generation in water areas,'' \emph{Remote Sensing}, vol.~12, no.~5, p. 870, 2020.

\bibitem{barnes2009patchmatch}
C.~Barnes, E.~Shechtman, A.~Finkelstein, and D.~B. Goldman, ``Patchmatch: A randomized correspondence algorithm for structural image editing,'' \emph{ACM Trans. Graph.}, vol.~28, no.~3, p.~24, 2009.

\bibitem{he2019learing}
H.~He, M.~Chen, T.~Chen, D.~Li, and P.~Cheng, ``Learning to match multitemporal optical satellite images using multi-support-patches siamese networks,'' \emph{Remote Sensing Letters}, vol.~10, no.~6, pp. 516--525, 2019.

\bibitem{bleyer2011patchmatch}
M.~Bleyer, C.~Rhemann, and C.~Rother, ``Patchmatch stereo-stereo matching with slanted support windows.'' in \emph{Bmvc}, vol.~11, 2011, pp. 1--11.

\bibitem{franchis2014pipeline}
C.~de~Franchis, E.~Meinhardt-Llopis, J.~Michel, J.-M. Morel, and G.~Facciolo, ``An automatic and modular stereo pipeline for pushbroom images,'' \emph{ISPRS Annals of the Photogrammetry, Remote Sensing and Spatial Information Sciences}, vol. II-3, pp. 49--56, 2014.

\bibitem{franchis2014refine}
{{de Franchis, Carlo and Meinhardt-Llopis, Enric and Michel, Julien and Morel, J-M and Facciolo, Gabriele}}, ``Automatic sensor orientation refinement of pl{\'e}iades stereo images,'' in \emph{2014 IEEE Geoscience and Remote Sensing Symposium}.\hskip 1em plus 0.5em minus 0.4em\relax IEEE, 2014, pp. 1639--1642.

\bibitem{franchis2014rect}
{de Franchis, Carlo and Meinhardt-Llopis, Enric and Michel, Julien and Morel, J-M and Facciolo, Gabriele}, ``On stereo-rectification of pushbroom images,'' in \emph{2014 IEEE International Conference on Image Processing (ICIP)}.\hskip 1em plus 0.5em minus 0.4em\relax IEEE, 2014, pp. 5447--5451.

\bibitem{chang2018pyramid}
J.-R. Chang and Y.-S. Chen, ``Pyramid stereo matching network,'' in \emph{Proceedings of the IEEE Conference on Computer Vision and Pattern Recognition}, 2018, pp. 5410--5418.

\bibitem{yang2019hsm}
G.~Yang, J.~Manela, M.~Happold, and D.~Ramanan, ``Hierarchical deep stereo matching on high-resolution images,'' in \emph{The IEEE Conference on Computer Vision and Pattern Recognition (CVPR)}, June 2019.

\bibitem{he2022hmsmnet}
S.~He, S.~Li, S.~Jiang, and W.~Jiang, ``Hmsm-net: Hierarchical multi-scale matching network for disparity estimation of high-resolution satellite stereo images,'' \emph{ISPRS Journal of Photogrammetry and Remote Sensing}, vol. 188, pp. 314--330, 06 2022.

\bibitem{kaizhang2020}
K.~Zhang, G.~Riegler, N.~Snavely, and V.~Koltun, ``Nerf++: Analyzing and improving neural radiance fields,'' \emph{arXiv:2010.07492}, 2020.

\bibitem{yu2022plenoxel}
S.~Fridovich-Keil, A.~Yu, M.~Tancik, Q.~Chen, B.~Recht, and A.~Kanazawa, ``Plenoxels: Radiance fields without neural networks,'' in \emph{2022 IEEE/CVF Conference on Computer Vision and Pattern Recognition (CVPR)}, 2022, pp. 5491--5500.

\bibitem{sun2022dvgo}
C.~Sun, M.~Sun, and H.-T. Chen, ``Direct voxel grid optimization: Super-fast convergence for radiance fields reconstruction,'' in \emph{2022 IEEE/CVF Conference on Computer Vision and Pattern Recognition (CVPR)}, 2022, pp. 5449--5459.

\bibitem{yariv2020multiview}
L.~Yariv, Y.~Kasten, D.~Moran, M.~Galun, M.~Atzmon, B.~Ronen, and Y.~Lipman, ``Multiview neural surface reconstruction by disentangling geometry and appearance,'' \emph{Advances in Neural Information Processing Systems}, vol.~33, 2020.

\bibitem{Niemeyer2019DifferentiableVR}
M.~Niemeyer, L.~M. Mescheder, M.~Oechsle, and A.~Geiger, ``Differentiable volumetric rendering: Learning implicit 3d representations without 3d supervision,'' \emph{2020 IEEE/CVF Conference on Computer Vision and Pattern Recognition (CVPR)}, pp. 3501--3512, 2019.

\bibitem{Oechsle2021unisurf}
M.~Oechsle, S.~Peng, and A.~Geiger, ``Unisurf: Unifying neural implicit surfaces and radiance fields for multi-view reconstruction,'' in \emph{International Conference on Computer Vision (ICCV)}, 2021.

\bibitem{yariv2021volsdf}
L.~Yariv, J.~Gu, Y.~Kasten, and Y.~Lipman, ``Volume rendering of neural implicit surfaces,'' in \emph{Proceedings of the 35th International Conference on Neural Information Processing Systems}, ser. NIPS '21.\hskip 1em plus 0.5em minus 0.4em\relax Red Hook, NY, USA: Curran Associates Inc., 2024.

\bibitem{Sun2022Neural3R}
J.~Sun, X.~Chen, Q.~Wang, Z.~Li, H.~Averbuch-Elor, X.~Zhou, and N.~Snavely, ``Neural 3d reconstruction in the wild,'' \emph{ACM SIGGRAPH 2022 Conference Proceedings}, 2022.

\bibitem{li2023neuralangelo}
Z.~Li, T.~M\"uller, A.~Evans, R.~H. Taylor, M.~Unberath, M.-Y. Liu, and C.-H. Lin, ``Neuralangelo: High-fidelity neural surface reconstruction,'' in \emph{IEEE Conference on Computer Vision and Pattern Recognition ({CVPR})}, 2023.

\bibitem{wu2022voxurf}
T.~Wu, J.~Wang, X.~Pan, X.~Xu, C.~Theobalt, Z.~Liu, and D.~Lin, ``Voxurf: Voxel-based efficient and accurate neural surface reconstruction,'' in \emph{International Conference on Learning Representations (ICLR)}, 2023.

\bibitem{behari2024sundial}
N.~Behari, A.~Dave, K.~Tiwary, W.~Yang, and R.~Raskar, ``Sundial: 3d satellite understanding through direct, ambient, and complex lighting decomposition,'' in \emph{2024 IEEE/CVF Conference on Computer Vision and Pattern Recognition Workshops (CVPRW)}.\hskip 1em plus 0.5em minus 0.4em\relax Los Alamitos, CA, USA: IEEE Computer Society, jun 2024, pp. 522--532.

\bibitem{qu_sat-mesh_2023}
Y.~Qu and F.~Deng, ``\BIBforeignlanguage{en}{Sat-{Mesh}: {Learning} {Neural} {Implicit} {Surfaces} for {Multi}-{View} {Satellite} {Reconstruction}},'' \emph{\BIBforeignlanguage{en}{Remote Sensing}}, vol.~15, no.~17, p. 4297, Jan. 2023.

\bibitem{billouard2024satngp}
C.~Billouard, D.~Derksen, E.~Sarrazin, and B.~Vallet, ``Sat-ngp: Unleashing neural graphics primitives for fast relightable transient-free 3d reconstruction from satellite imagery,'' \emph{arXiv preprint arXiv:2403.18711}, 2024.

\bibitem{max1995optical}
N.~Max, ``Optical models for direct volume rendering,'' \emph{IEEE Transactions on Visualization and Computer Graphics}, vol.~1, no.~2, pp. 99--108, 1995.

\bibitem{dfc2019supp1}
G.~Christie, K.~Foster, S.~Hagstrom, G.~D. Hager, and M.~Z. Brown, ``Single view geocentric pose in the wild,'' in \emph{2021 IEEE/CVF Conference on Computer Vision and Pattern Recognition Workshops (CVPRW)}, 2021, pp. 1162--1171.

\bibitem{dfc2019supp2}
G.~Christie, R.~R. R.~M. Abujder, K.~Foster, S.~Hagstrom, G.~D. Hager, and M.~Z. Brown, ``Learning geocentric object pose in oblique monocular images,'' in \emph{Proceedings of the IEEE/CVF Conference on Computer Vision and Pattern Recognition}, 2020, pp. 14\,512--14\,520.

\bibitem{VisSat-2019}
K.~Zhang, J.~Sun, and N.~Snavely, ``Leveraging vision reconstruction pipelines for satellite imagery,'' in \emph{ICCV Workshop on 3D Reconstruction in the Wild (3DRW)}, 2019.

\bibitem{zhang2023spsnerf}
L.~Zhang and E.~Rupnik, ``Sparsesat-nerf: Dense depth supervised neural radiance fields for sparse satellite images,'' \emph{ISPRS Annals}, 2023.

\end{thebibliography}

\vfill

\end{document}